\documentclass{ieeeaccess}
\usepackage{cite}
\usepackage{amsmath,amssymb,amsfonts}
\usepackage{algorithmic}
\usepackage{graphicx}
\usepackage{textcomp}

\usepackage{color}
\usepackage{tabularray}
\usepackage{hhline}
\usepackage{colortbl}

\usepackage{footnote}

\usepackage{siunitx}
\usepackage{url}

\usepackage{bm}
\makeatletter
\AtBeginDocument{\DeclareMathVersion{bold}
\SetSymbolFont{operators}{bold}{T1}{times}{b}{n}
\SetSymbolFont{NewLetters}{bold}{T1}{times}{b}{it}
\SetMathAlphabet{\mathrm}{bold}{T1}{times}{b}{n}
\SetMathAlphabet{\mathit}{bold}{T1}{times}{b}{it}
\SetMathAlphabet{\mathbf}{bold}{T1}{times}{b}{n}
\SetMathAlphabet{\mathtt}{bold}{OT1}{pcr}{b}{n}
\SetSymbolFont{symbols}{bold}{OMS}{cmsy}{b}{n}
\renewcommand\boldmath{\@nomath\boldmath\mathversion{bold}}}
\makeatother

\def\BibTeX{{\rm B\kern-.05em{\sc i\kern-.025em b}\kern-.08em
    T\kern-.1667em\lower.7ex\hbox{E}\kern-.125emX}}

\begin{document}
\pagestyle{plain}

% \history{Date of publication xxxx 00, 0000, date of current version xxxx 00, 0000.}
\doi{10.1109/ACCESS.2024.0429000}

\title{Efficient Frame Extraction: A Novel Approach Through Frame Similarity and Surgical Tool Tracking for Video Segmentation}
\author{
\uppercase{Huu Phong Nguyen}\authorrefmark{1}, 
\uppercase{Shekhar Madhav Khairnar}\authorrefmark{1},
\uppercase{Sofia Garces Palacios}\authorrefmark{1},
\uppercase{Amr Al-Abbas}\authorrefmark{1},
\uppercase{Melissa E. Hogg}\authorrefmark{2},
\uppercase{Amer H. Zureikat}\authorrefmark{3},
\uppercase{Patricio M. Polanco}\authorrefmark{1},
\uppercase{Herbert Zeh III}\authorrefmark{1},
\uppercase{Ganesh Sankaranarayanan\authorrefmark{1}}}
\address[1]{Department of Surgery, University of Texas Southwestern Medical Center, Texas, USA}
\address[2]{NorthShore University HealthSystem, Evanston, IL, USA}
\address[3]{University of Pittsburgh Medical Center, Pittsburgh, PA, USA}
% \tfootnote{This paragraph of the first footnote will contain support
% information, including sponsor and financial support acknowledgment. For
% example, ``This work was supported in part by the U.S. Department of
% Commerce under Grant BS123456.''}
% \markboth
% {Author \headeretal: Preparation of Papers for IEEE TRANSACTIONS and JOURNALS}
% {Author \headeretal: Preparation of Papers for IEEE TRANSACTIONS and JOURNALS}

\corresp{Corresponding authors: Huu Phong Nguyen (e-mail: huuphong.nguyen@utsouthwestern.edu) and Ganesh Sankaranarayanan(e-mail: ganesh.sankaranarayanan@utsouthwestern.edu).}

\begin{abstract}
The interest in leveraging Artificial Intelligence (AI) for surgical procedures to automate analysis has witnessed a significant surge in recent years. One of the primary tools for recording surgical procedures and conducting subsequent analyses, such as performance assessment, is through videos. However, these operative videos tend to be notably lengthy compared to other fields, spanning from thirty minutes to several hours, which poses a challenge for AI models to effectively learn from them. Despite this challenge, the foreseeable increase in the volume of such videos in the near future necessitates the development and implementation of innovative techniques to tackle this issue effectively. In this article, we propose a novel technique called Kinematics Adaptive Frame Recognition (KAFR) that can efficiently eliminate redundant frames to reduce dataset size and computation time while retaining useful frames to improve accuracy. Specifically, we compute the similarity between consecutive frames by tracking the movement of surgical tools. Our approach follows these steps: $i)$ Tracking phase: a YOLOv8 model is utilized to detect tools presented in the scene, $ii)$ Similarity phase: Similarities between consecutive frames are computed by estimating variation in the spatial positions and velocities of the tools, $iii$) Classification phase: An X3D CNN is trained to classify segmentation. We evaluate the effectiveness of our approach by analyzing datasets obtained through retrospective reviews of cases at two referral centers. The newly annotated Gastrojejunostomy (GJ) dataset covers procedures performed between 2017 and 2021, while the previously annotated Pancreaticojejunostomy (PJ) dataset spans from 2011 to 2022 at the same centers. In the GJ dataset, each robotic GJ video is segmented into six distinct phases. By adaptively selecting relevant frames, we achieve a \textbf{tenfold} reduction in the number of frames while improving \textbf{accuracy} by $4.32\%$ (from 0.749 to 0.7814) and the F1 score by $0.16\%$. Our approach is also evaluated on the PJ dataset, demonstrating its efficacy with a fivefold reduction of data and a $2.05\%$ accuracy improvement (from 0.8801 to 0.8982), along with $2.54\%$ increase in F1 score (from 0.8534 to 0.8751). In addition, we also compare our approach with the state-of-the-art approaches to highlight its competitiveness in terms of performance and efficiency. Although we examined our approach on the GJ and PJ datasets for phase segmentation, this could also be applied to broader, more general datasets. Furthermore, KAFR can serve as a supplement to existing approaches, enhancing their performance by reducing redundant data while retaining key information, making it a valuable addition to other AI models.
\end{abstract}


\titlepgskip=-21pt

\maketitle

\section{Introduction}
\label{sec:introduction}
\IEEEPARstart{T}{h}e growing application of AI in video analysis is primarily inspired by developments in modern Graphics Processing Units (GPUs)~\cite{phong2022pso}. Traditional computer vision-based video analysis for applications such as activity recognition, video summary, etc., faced challenges due to the higher dimensional nature of the data and the need to track both spatial and temporal features~\cite{nguyen2023video,al2024development}. However, the emergence of AI has enabled researchers to develop increasingly intricate and data-driven approaches for interpreting video. This has led to considerable performance increases for a variety of applications, for example, object detection~\cite{liu2023yolo,liang2015recurrent,felzenszwalb2009object}, action recognition~\cite{sun2024k,phong2023pattern}, and video segmentation~\cite{garg2024human,grammatikopoulou2024spatio}. AI models can effectively learn complex patterns and temporal correlations directly from raw video data, revolutionizing the field and opening up new opportunities for applications in surveillance~\cite{li2020abnormal}, autonomous vehicles~\cite{gao2024human}, and healthcare~\cite{IJMESD400}.

In the field of surgery, the use of AI in analyzing surgical videos has a great potential~\cite{josiah2024artificial,knudsen2024clinical,lavanchy2023preserving}, including automating activity recognition from video recordings~\cite{hegde2024automated,demir2023deep}. However, the duration of surgery videos might span from thirty minutes to several hours, posing a significant challenge for AI models, particularly in processing and analyzing lengthy sequential data effectively.

Early approaches to handling this challenge leveraged statistical models designed for sequential data, such as conditional random fields~\cite{quellec2014real} and hidden Markov models~\cite{padoy2008line}. However, due to the complex temporal relations among frames, these techniques exhibit limited representation capacities with predefined dependencies~\cite{jin2017sv}. 

Nonetheless, as deep learning technology evolved and gained prominence in the research community, a plethora of novel techniques have emerged, achieving impressive outcomes. When dealing with sequential data, models such as Gated Recurrent Unit (GRU)~\cite{valipour2017recurrent}, integration of Long Shot-Term Memory (LSTM) and Convolutional Neural Networks (CNN)~\cite{jalal2021deep}, and bi-directional LSTMs can be utilized~\cite{he2021db}.

LSTM models, however, encounter the vanishing gradient problem when faced with long sequences of data. Alternatively, 3-D CNN models~\cite{al2024development,hegde2024automated} are favored for processing sequential data. This preference stems from their capacity to capture both spatial and temporal information simultaneously. A few notable examples illustrate this approach, including a 3-D fully CNN (3DFCNN) that automatically encodes spatio-temporal patterns from these sequences, enabling action classification based on both spatial and temporal information. The method presents an end-to-end solution for human action recognition directly from raw depth image-sequences~\cite{sanchez20223dfcnn}.
% Fractioned Adjacent Spatial and Temporal (FAST) 3-D convolutions are used to extract spatio-temporal patterns from 3-D videos, expanding on 2-D image convolutions~\cite{stergiou2019spatio}. 
Additionally, CNN models have been effective in distinguishing between normal and Alzheimer's disease patients utilizing an auto-encoder architecture and 3-D CNNs~\cite{hosseini2016alzheimer}. Furthermore, a fusion network that integrates the 3-D convolutional posture stream with the 2-D convolutional stream enhances the accuracy of identifying human actions~\cite{huang2018human}. However, compared to 2-D CNNs, 3-D CNNs can be more resource-intensive and challenging to train, especially when processing a large number of frames, due to their higher computational and memory needs.

Vision transformers~\cite{nguyen2023video,han2023flatten,liu2021swin,arnab2021vivit} can handle longer sequences than CNN-LSTM models, but they require a large amount of data to train efficiently, which is often scarce and expensive in surgery.

Since surgical videos are lengthy and complex, increasingly advanced techniques are required to manage the intricate temporal dependencies. To address this issue, our approach involves fine-tuning by identifying key frames and removing non-essential ones, thereby reducing the total number of frames for training. Focusing on these key frames might help to train the network more efficiently by lowering computational costs and speeding up the processing. Specifically, in this approach, i) background noise is filtered out while selectively preserving objects of interest, which are surgical tools in our work; ii) Adaptive 1 and Adaptive 2 methods are proposed to select key frames; iii) Distinct X3D CNN models are trained using the selected frames.

The primary contributions of this paper are:
\begin{enumerate}
    \item The introduction of the use of kinematics data (velocity and acceleration) with tool tracking for adaptively selecting key frames in video-based phase segmentation in surgery.
    \item A new and comprehensive  Gastrojejunostomy (GJ) dataset is collected and annotated.
    \item Extensive experiments on GJ and Pancreaticojejunostomy (PJ) datasets are conducted to validate the effectiveness of the proposed methods.
    \item The source code is made available on GitHub (github.com) to facilitate reproduction of this work\footnote{\url{https://github.com/leonlha/Kinematics-AFR}}.
\end{enumerate}

The remainder of the article is organized as follows: In Section~\ref{sec:related_works}, we discuss relevant approaches in applying Deep Learning in Surgery and selecting key frames. In addition, the proposed methods, namely Adaptive 1 and Adaptive 2, are introduced in Section~\ref{sec:proposed}. The results of experiments and the extension of the experiments are presented in Sections~\ref{sec:result}, and~\ref{sec:extension}. Moreover, the comparison with state-of-the-art methods is discussed in Section~\ref{sec:stateoftheart}, followed by discussions in Section~\ref{sec:discussions}. Finally, we conclude our work in Section~\ref{sec:conclusion}.
\section{Related Works}
\label{sec:related_works}
\subsection{Surgery Datasets}
Despite the growing interest in studying surgical skill performance, the number of public datasets remains limited. Unlike datasets in other fields, e.g., computer science, surgical datasets often contain sensitive patient information that must be kept confidential. Nevertheless, a few open datasets are available for research (Table~\ref{tab:table_datasets}).

One of the earliest surgical datasets, Cholec80, contains 80 videos of laparoscopic cholecystectomy surgeries performed by 13 surgeons. The dataset is divided into a training set (40 videos) and a testing set (40 videos), with the videos segmented into 7 phases. A subset of Cholec80, the Cholec51 dataset, includes 51 videos. However, our datasets differ as they focus on the reconstruction portion of the robotic pancreaticodudenectomy procedure, namely, the Gastrojejunostomy and Pancreaticojejunostomy anastomoses.

Additional datasets include the M2CAI16 Workflow Challenge, which consists of 41 laparoscopic cholecystectomy videos (27 for training and 14 for testing), segmented into 8 phases. The CATARACTS dataset, with 50 videos, features 19 phase categories. We collected a similar number of videos (GJ with 42 videos and PJ with 100 videos).

The JHU-ISI Gesture and Skill Assessment Working Set (JIGSAWS) contains 39 videos performed by 8 users, with each video recorded approximately five times. While JIGSAWS is frequently used in surgical skill assessment research, it focuses on simulation tasks, whereas our datasets were recorded in operating rooms during actual surgeries.
\definecolor{Silver}{rgb}{0.752,0.752,0.752}
\begin{table}[!t]
\centering
\caption{List of surgery datasets.}
\label{tab:table_datasets}
\begin{tblr}{
  row{1} = {Silver},
  column{2} = {c},
  column{3} = {c},
}
\textbf{Name} & \textbf{Videos} & \textbf{Phases} & \textbf{Description}                                                        \\
Cholec80~\cite{twinanda2016endonet}              & 80             & 7                         & {Cholecystectomy procedures\\from INRIA}                                    \\
Cholec51~\cite{twinanda2016endonet}              & 51             & 7                         & A subset of Cholec80                                                      \\
{CATARACTS\\~\cite{ALHAJJ201924}}             & 50             & 19                        & {Cataract procedures from\\INRIA}                                         \\
M2cai16~\cite{cadene2016m2cai}               & 41             & 8                         & {Minimally invasive surgical\\procedures from the \\M2CAI 2016 Challenge}   \\
JIGSAWS~\cite{gao2014jhu}               & 17             & 4                         & {Simulation procedures from\\JHU}\\
PJ~\cite{al2024development}               & 100             & 6                         & {Pancreaticojejunostomy}\\
\hline
\textbf{GJ}               & \textbf{42}             & \textbf{6}                         & {\textbf{Gastrojejunostomy}\\\textbf{(this work)}}
\end{tblr}
\end{table}

The PJ dataset, which we collected previously, comprises 100 videos segmented into 6 phases of Pancreaticojejunostomy. We use this dataset to provide a comparative baseline for our current research.
\subsection{Key Frame Recognition}
As mentioned in the previous section, surgical videos tend to be longer than those in other fields, which poses a challenge for effectively using deep learning methods. A non-trivial approach entails significantly reducing the number of frames, for example, through Uniform Frame Sampling (UFS), which skips frames at regular intervals. Early attempts for key frame extraction relied on shot boundary-based algorithms~\cite{boreczky1996comparison}. Essentially, this technique utilizes the first or middle frame of each shot as the key frame after shot boundary detection. Although shot boundary-based approaches are simple to apply and generalize, the extracted key frame cannot fully capture the visual content. Another example used a Long Short-Term Memory (LSTM) network linked to the output of the underlying CNN~\cite{donahue2015long}. However, this framework obtains sixteen sample frames evenly split with an eight-frame stride from the full-length videos as the video representation. Although straightforward, this technique disregards the significance of consecutive frames, implying that certain frames hold significantly more importance than others.

One solution involves computing the similarity between frames and discarding those that exceed a certain threshold. For instance, according to Savran et al.~\cite{savran2023novel}, frames are compared using measures like the Mean Squared Error (MSE) on human activity datasets, i.e., KTH and UCF-101. When high similarity between consecutive frames is observed, indicating minimal changes in the action, these frames are considered redundant and removed. By discarding frames that exceed a predefined similarity threshold, the technique effectively filters out frames that do not provide new or significant information to the video’s action. However, strategies that are effective in other fields may not be applicable in surgery. For example, tools for minimally invasive surgery, such as scissors and forceps (which need to be tracked), often occupy a small portion of the screen and are overshadowed by background noise. It is worth noting that human organs are constantly in motion, presenting significant challenges that render approaches like background subtraction, optical flow, and even depth estimation, as well as non-deep learning methods such as MSE and Structural Similarity Index (SSIM) (which use all pixels of the image), less efficient~\cite{savran2023novel,yang2021fast,zhong2020key,cucchiara2003detecting}.

\section{Proposed Methods}
\label{sec:proposed}
\subsection{Architecture}
\begin{figure} [bt!]
\begin{center}
\includegraphics[keepaspectratio,width=0.45\textwidth]{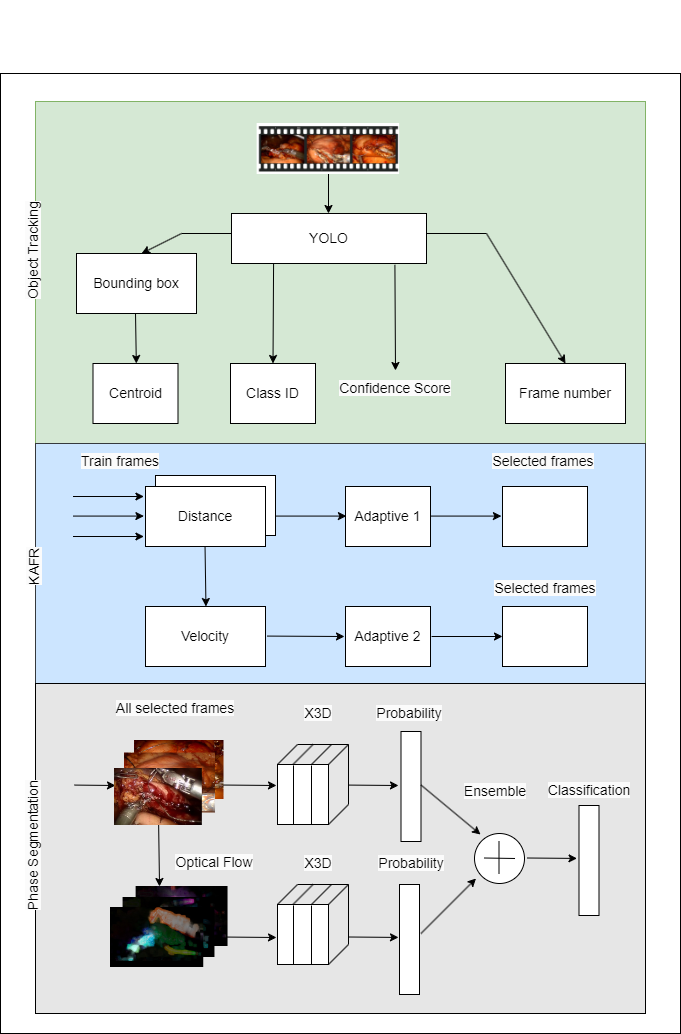}
\caption{Overview of the KAFR architecture. It involves three phases: (1) Object Tracking -- Detects and tracks surgical tools. (2) KAFR -- Computes centroids and identifies critical frames using tool movement. (3) Phase Segmentation -- Classifies frames into segments with two X3D CNN models in a two-stream convolutional networks configuration.}
\label{fig:architecture_m}
\end{center}
\end{figure}

Figure~\ref{fig:architecture_m} presents an overview of our proposed architecture for KAFR. This workflow is divided into three distinct steps: Object Tracking, KAFR, and Phase Segmentation. The first phase -- Object Tracking -- plays an important role because the module traces the locations of surgical tools, which is critical for computing KAFR, especially when the tools are surrounded by a noisy environment. In this phase, we extract the bounding box, class ID (with a confidence score greater than $0.5$), and frame number from the input video for surgical instruments. In the second phase, we compute the \textbf{centroid} of each tool, enabling us to track its movement. Please note that only frames from the training set were chosen. With the tools detected as polygon shapes, the centroids were determined using the GeoPandas library in PyTorch. The distance and velocity are utilized as indicators to detect whether the surgeon's maneuvers are essential. Subsequently, critical frames are identified. In the third phase, we employ two distinct X3D CNN models to classify frames into six different segments.
\subsection{Kinematics Adaptive Frame Recognition}
\subsubsection{Adaptive 1}
In a video sequence of $n$ frames $\{x_1, x_2, ..., x_n\}$, a threshold $d$ determines the distance measure $D(x_i, x_j)$ used to classify a pair of frames $(x_i, x_j)$ as key frames or similar frames. Key frame pairs are denoted as $P_{\text{key}}$. Frames $x_i$ and $x_j$ are considered key frames if their distance is less than or equal to $d$. The frames between the two key frames are similar and denoted by $P_{\text{similar}}$. The search for key frame pairs starts from the first frame until the last frame is encountered. Thus, we define a set of all key (K) pairs as follows:
\begin{equation}
K(d) = \{(x_i,x_j) \mid D(x_i,x_j) \leq d, (x_i,x_j) \in P_{key}\},
\label{eq:f1}
\end{equation}
Assuming $S$ is a set of points (one pixel for each point) of a frame and $s$ is a subset of $S$, the distance $D(x_i, x_j)$ is defined as
\begin{equation}
D(x_i, x_j) = f\left(\sum\limits_{s \in S}\sum_{k=i+1}^{j}\left|\left|s(x_i)-s(x_k)\right|\right|\right),
\end{equation}
with $\left|\left|\cdot\right|\right|$ being the Euclidean norm and $f\,:\,\mathbb{R}\rightarrow \mathbb{R}$ being a decreasing (or at least non-increasing) function. In \underline{Adaptive 1}, we assume that
\begin{equation}
f(z_d)= \frac{1}{\left(z_d+\epsilon\right)^{\beta_d}},
\end{equation}
for constant $\beta_d>0$ and $\epsilon$ is a small number introduced to avoid division by zero.
\begin{figure} [tbh!]
\begin{center}
\includegraphics[keepaspectratio,width=0.45\textwidth]{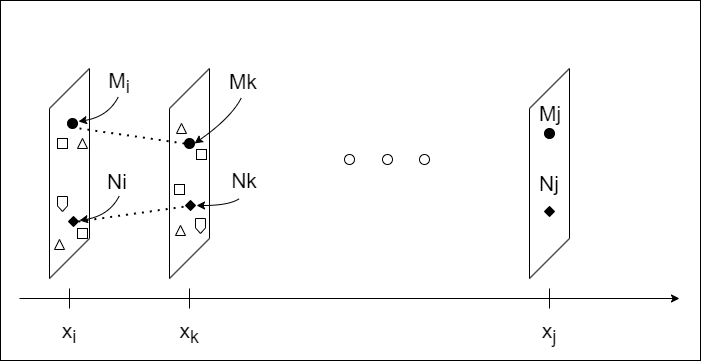}
\caption{Illustration of Adaptive Frame Recognition. The subset $s$ consists of two tools, $M$ and $N$ (represented as circle and diamond shapes), emerging from a noisy environment (which might include organs of the abdominal cavity, other surgical tools, and noise, represented by other shapes). The tools are tracked across frames. The frame index $k$ is incremented one at a time until a designated threshold $d$ is reached.}
\label{fig:image1}
\end{center}
\end{figure}
\subsubsection{Adaptive 2}
One alternative to the equation~\eqref{eq:f1} is using variation of \emph{velocity} rather than distance
\begin{equation}
K(d) = \{(x_i,x_j) \mid V(x_i,x_j) \leq d, (x_i,x_j) \in P_{key}\},
\end{equation}

Assuming $S$ is a set of points of a frame and $s$ is a subset $S$, the variation of velocity $V(x_i, x_j)$ is denoted as
\begin{equation}
V(x_i, x_j) = f\left( \sum\limits_{\substack{s \in S}} \sum_{k=i+1}^{j} \left| V_{\text{s}}(x_i) - V_{\text{s}}(x_k) \right| \right),
\end{equation}
where $V_{\text{s}}(x_k)$ is the velocity of point {$x_k$} in the subset {s}

In \underline{Adaptive 2}, we assume that
\begin{equation}
f(z_v)= \frac{1}{\left(z_v+\epsilon\right)^{\beta_v}}.
\end{equation}
for constant $\beta_v>0$ and $\epsilon$ is a small number introduced to avoid division by zero.
\subsection{Object Detection}
\label{sec:objectdetection}
\begin{figure} [b!]
\begin{center}
\includegraphics[keepaspectratio,width=0.45\textwidth]{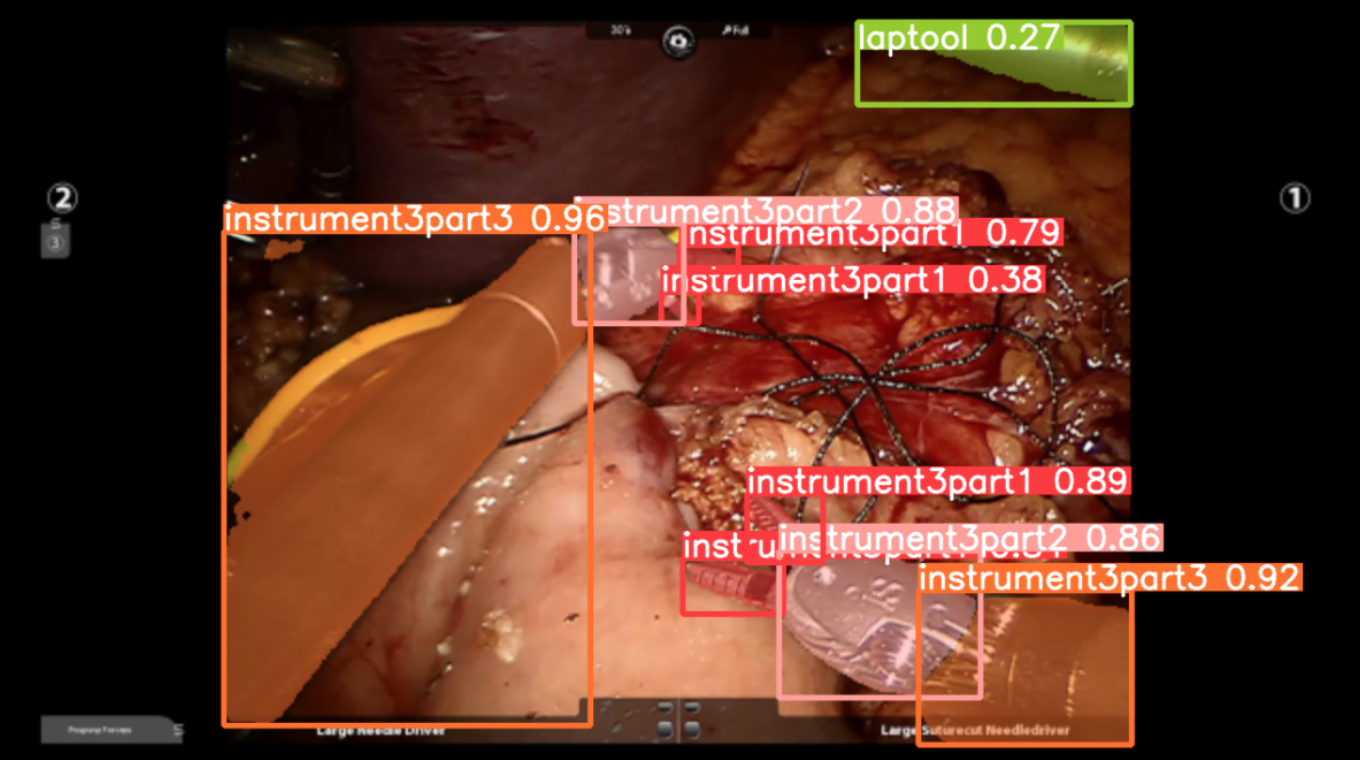}
\caption{Tools auto-detection using Yolo.}
\label{fig:yolo}
\end{center}
\end{figure}

As mentioned earlier, the detection of surgical tools is crucial. Therefore, we propose using the YOLOv8 neural network, the latest advancement in the ``You Only Look Once'' (YOLO) framework~\cite{redmon2016you}. It is well-known for its real-time object detection capabilities, outperforming those of fast and faster Region CNN (RCNN)~\cite{girshick2014rich} for tool detection (Figure~\ref{fig:yolo}). While YOLO comes pre-trained on general datasets like COCO~\cite{lin2014microsoft} or ImageNet~\cite{deng2009imagenet}, its direct application in surgical video analysis may be suboptimal. Surgical instruments and operating environments possess unique characteristics, necessitating fine-tuning of the model for optimal performance.

\begin{figure} [tbp!]
\begin{center}
\includegraphics[keepaspectratio,width=0.45\textwidth]{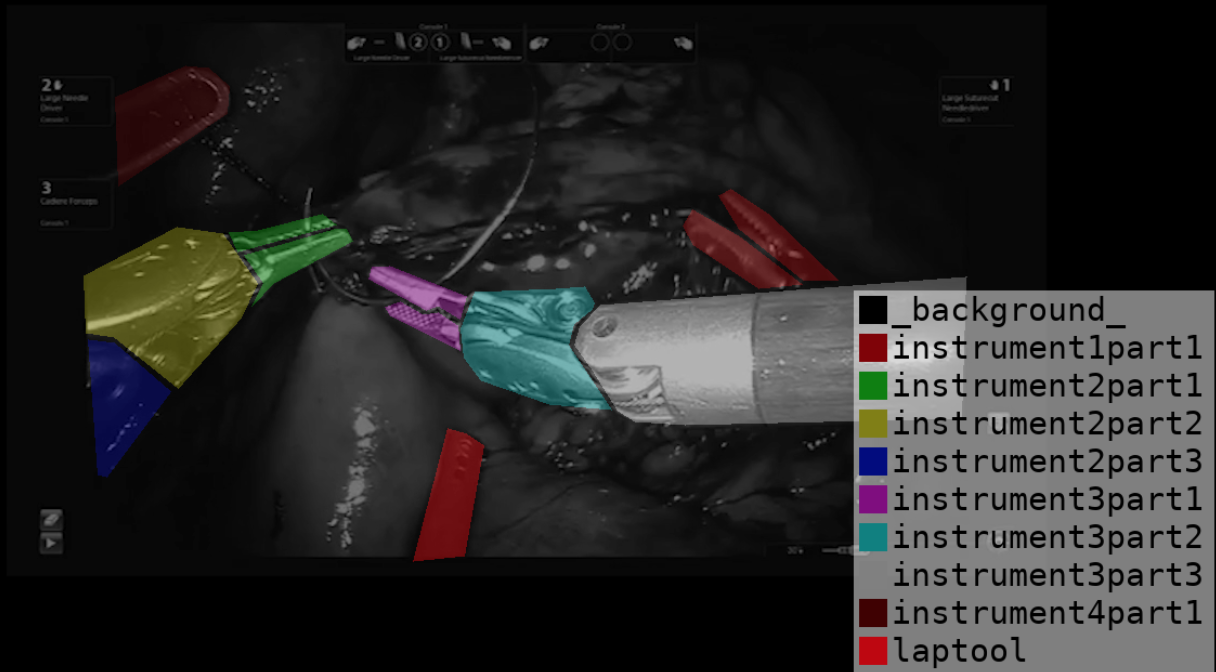}
\caption{LabelMe for surgical tools annotation.}
\label{fig:labelme}
\end{center}
\end{figure}

We used a dataset comprising 28 surgical videos and employed the LabelMe~\cite{russell2008labelme} tool to generate annotations, yielding 917 annotated images (Figure~\ref{fig:labelme}). To the best of our knowledge, there were not any other datasets containing the same set of surgical tools as those in the GJ and PJ datasets. Therefore, we used a portion of the GJ videos for tool segmentation. Although the same GJ dataset was used for both tasks--recognizing individual surgical tools and phase segmentation--these tasks are fundamentally different and should not interfere with each other’s performance. 

In addition, LabelMe automatically labels any part of an image that is not explicitly masked by the user as background. This classification is crucial for training detection models because it trains them to distinguish between objects of interest and irrelevant areas.

YOLO framework splits the input image into a grid of $S$ rows and $S$ columns, where $S$ is a desired number. Each cell generates candidate bounding boxes and confidence levels for the presence of instruments or tools. The confidence metric \( \text{Conf(Tool)} \) is determined by the binary indicator \( \text{Pr(Tool)} \) and is defined as:

\begin{equation}
\text{Conf(Tool)} = \text{Pr(Tool)} \times \text{IOU}_{\text{pred}}^{\text{truth}}
\end{equation}
where \( \text{Pr(Tool)} \) denotes the presence of a tool (0 for absent, 1 for present). The Intersection Over Union (IOU) between the predicted box and the true box is calculated as:
\begin{equation}
\text{IOU} = \frac{\text{area}(box_{\text{Truth}} \cap box_{\text{Pred}})}{\text{area}(box_{\text{Truth}} \cup box_{\text{Pred}})} 
\end{equation}

For more information on automated tool tracking, please refer to our other study~\cite{shekhar2024machine}. 
\subsection{Surgical Phase Segmentation}
\subsubsection{X3D Convolutional Neural Networks}
\begin{figure*}[tb!]
\begin{center}
\includegraphics[keepaspectratio,width=0.95\textwidth]{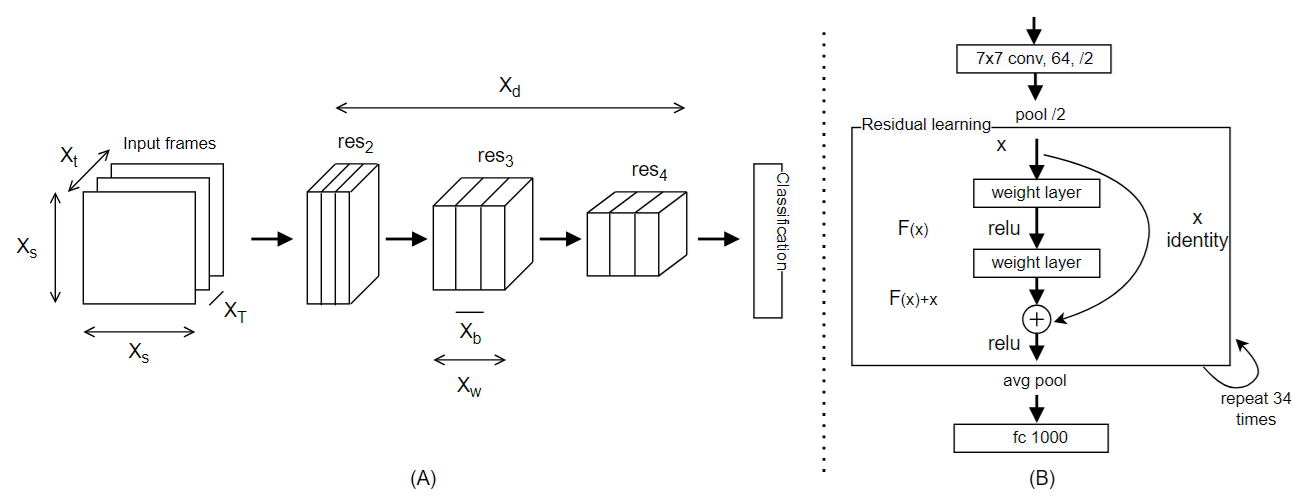}
\caption{Render X3D model architecture. (A) The X3D network extends a 2-D ResNet (res) one axis at a time across various axes, including temporal duration $X_t$, frame rate $X_T$, spatial resolution $X_s$, width $X_w$, bottleneck width $X_b$, and depth $X_d$. (B) Details of the 2-D ResNet demonstrate residual learning (Adapted from~\cite{feichtenhofer2020x3d} and~\cite{he2016deep}).}
\label{fig:x3d}
\end{center}
\end{figure*}
Surgical phase segmentation is an important aspect of computer-assisted surgery and medical image analysis involving the automated partitioning of surgical procedures into distinct phases or stages using video or image data~\cite{demir2023deep, funke2023metrics, guedon2021deep, touma2022development, meireles2021sages, golany2022artificial}. The primary goal of surgical phase segmentation is to comprehend the temporal progression of surgeries, facilitating various clinical applications such as workflow analysis and skill assessment~\cite{garrow2021machine,tran2017phase}.

To identify these phases, we utilized the X3D architecture, a highly efficient 3-D CNN designed for activity recognition and video classification (Figure~\ref{fig:x3d}). X3D is an extension of the 2-D ResNet model originally designed for image classification, using residual learning to preserve essential information as it propagates through layers~\cite{feichtenhofer2020x3d,he2016deep,pydimarry2024evaluating}.

Optimized source code from prior research was reused~\cite{al2024development}, in which the model was enhanced through a variety of advanced techniques. To facilitate reference, these techniques are briefly described as follows. First, a re-sampling approach was applied: shorter steps were oversampled and longer steps undersampled, yielding a desired number of frames per step. This created a balanced frame distribution aligned with the median step duration. Second, a timestamp feature was added to encode the position of each frame with sinusoidal positional encoding. This generated a position embedding vector matching the size of the final X3D layer and embedded temporal context into the feature space. Next, the model was trained with a combined loss function: cross-entropy and Earth Mover’s Distance (EMD). EMD, which measures the effort to match two distributions, was used to reflect the sequential nature of the task, encouraging the model to relate temporally adjacent steps more closely. Finally, temporal smoothing was applied using a moving window of 31 frames (15 seconds before and after). The most frequent class within the window was utilized to update predictions, reducing short-term misclassifications.

\begin{figure}[tb!]
\begin{center}
\includegraphics[keepaspectratio,width=0.45\textwidth]{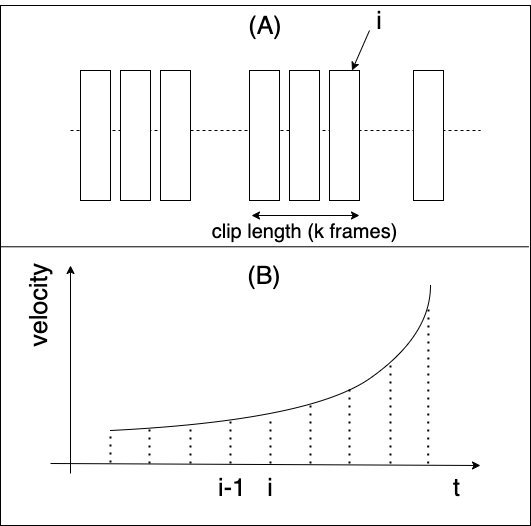}
\caption{Distinction between previous and proposed approaches. (A) Frame-by-Frame Level Phase Segmentation. (B) Uniform Frame Sampling vs. Kinematics Adaptive Frame Recognition.}
\label{fig:frame_by_frame}
\end{center}
\end{figure}

The previous research~\cite{al2024development} trained the X3D model based on a frame-by-frame approach, where, at any given time (e.g., at time $t_i$), the model was fed with $k$ frames, starting from frame $i$ and moving backward $k-1$ indices, as illustrated in the upper part of Figure~\ref{fig:frame_by_frame}. In their approach, frames were sampled at a fixed time interval. However, this research adopts a different strategy: we select only the key frames that capture significant tool movements. The lower part of the figure illustrates this difference. For example, as the tool moves with increasing velocity, the previous approach selects frames without accounting for this critical information. In contrast, our method selects only the key frames, such as the beginning and end frames, which not only dramatically reduce the size of the dataset but also signify moments that are likely to be crucial for classifying the phase.
\subsubsection{Two-stream Convolutional Networks}
We integrated optical flow (OF) into our approach for video action recognition because of its critical role in capturing motion dynamics and temporal changes. Optical flow estimation involves computing angle and magnitude shifts for each pixel across consecutive video frames. The Farneback~\cite{farneback2003two} algorithm is a common technique for dense optical flow computation that utilizes polynomial expansion and iterative field estimation. By solving a quadratic polynomial equation between two consecutive frames \( I(x,y,t) \) and \( I(x+\Delta x,y+\Delta y,t+\Delta t) \), optical flow \( (u,v) \) at each pixel is estimated to obtain displacement fields.

In the Phase Segmentation module shown in Figure~\ref{fig:architecture_m}, we trained two distinct X3D models independently. Instead of producing class labels in the final layer, we modified it to output class probabilities. These probability outputs were then ensembled to generate the final classifications.
\section{Experiments}
\label{sec:experiments}
\subsection{Benchmark Dataset}
\label{sec:dataset}
% Pancreatic cancer is projected to become the second leading cause of cancer-related deaths in the USA within the next two to three decades~\cite{mizrahi2020pancreatic}. 
%Cancer is the second biggest cause of death in the United States, with pancreatic cancer being fourth in terms of fatalities and having the lowest 5-year survival rate at only 13\%~\cite{siegel2024cancer}. Pancreatic cancer is also growing by roughly 1\% each year in both men and women in the United States. In 2024, a projected \num{66440} adults in the USA will be diagnosed with pancreatic cancer (3.3\% of all new cancer cases), and more than \num{51750} individuals will die from the cancer (8.4\% of all cancer fatalities).

The dataset used in the work is obtained from Robotic Pancreaticoduodenectomy (also known as the Whipple procedure) ~\cite{vining2021robotic,finks2011trends, crist1987improved}, which is a key procedure for treating pancreatic cancer. It involves two main phases: Resection and Reconstruction. The Reconstruction phase includes three anastomoses: Hepaticojejunostomy, Pancreaticojejunostomy, and Gastrojejunostomy. In the following, we will discuss the GJ and PJ datasets that are used in this study in more detail.
\subsubsection{GJ Dataset}
\label{sec:gj_dataset}
\begin{figure} [hbtp!]
\begin{center}
\includegraphics[keepaspectratio,width=0.45\textwidth]{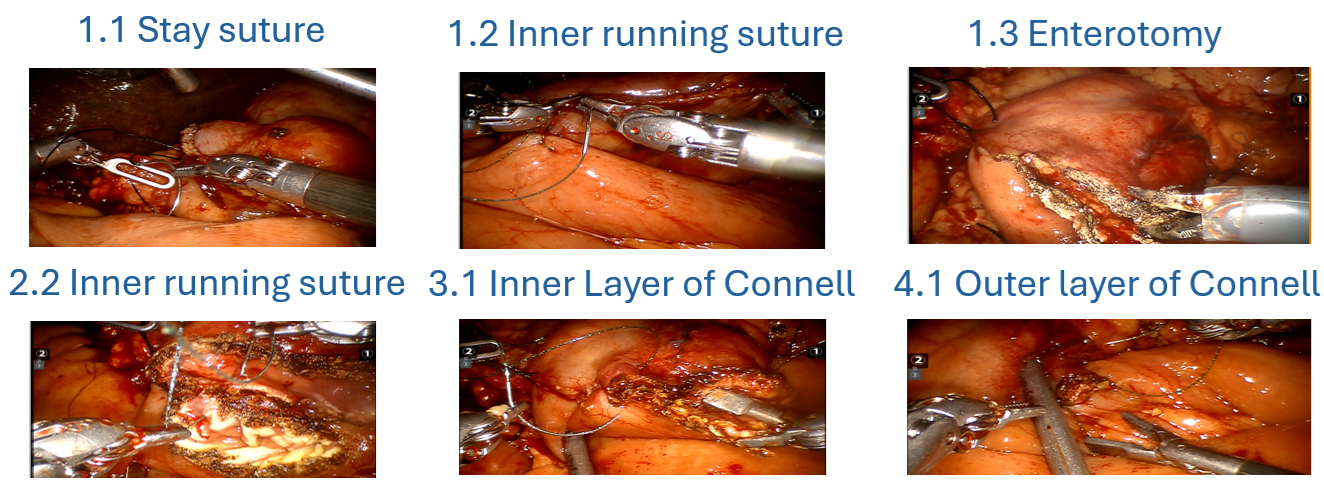}
\caption{Gastrojejunostomy Steps.}
\label{fig:gj_phases}
\end{center}
\end{figure}

A retrospective case review spanning from 2017 to 2021 was conducted at two prominent referral centers, UT Southwestern Medical Center and the University of Pittsburgh Medical Center. During this review, each robotic GJ video was segmented into six specific tasks: 1.1 Stay suture, 1.2 Inner running suture, 1.3 Enterotomy, 2.2 Inner running suture, 3.1 Inner Layer of Connell, and 4.1 Outer Layer of Connell, along with idle time. The idle phase, as defined in the literature, refers to the time characterized by the absence of clinically relevant motions~\cite{olsen2024surgical,quellec2014real}. We adhered to the same definitions when annotating the GJ dataset. The experts used a custom annotation framework to label these phases within the videos (Figure~\ref{fig:gj_phases}).
Of the 42 videos included, 30 were used for training and 12 for validation, with all frames annotated and extracted at the rate of 6 frames per second (fps), resulting in a dataset of \num{46903} images for training and \num{22647} images for validation. In the GJ dataset, videos may have been recorded at different times using different cameras, each configured with distinct frame rates (e.g., 24 and 30 fps). Since the videos are considerably long, the number of frames was reduced to shorten the training time. A rate of 6 was chosen, as it is the greatest common divisor of 24 and 30. For videos with other frame rates, we first converted them to 30 fps before selecting frames.
\subsubsection{PJ Dataset}
\label{sec:pj_dataset}
A retrospective review was conducted on recording videos collected from 2011 to 2022 across two quaternary referral centers: UT Southwestern Medical Center and the University of Pittsburgh Medical Center~\cite{al2024development}. Out of the 100 videos reviewed, 60 were utilized for model training, 10 for hyperparameter optimization, and 30 for performance testing. Frames were extracted at 6 frames/second and annotated. Each video was divided into tasks, with start and end times recorded for each. Six tasks were annotated per video: 1.1 Anterior Mattress Sutures, 1.2 Tying Mattress Sutures, 2.1 Enterotomy, 3.1 Posterior Duct Sutures, 3.2 Anterior Duct Sutures, and 4.1 Anterior Buttress Sutures.
\subsection{Distribution of Tools and Naming Conventions}
\begin{figure} [t!]
\begin{center}
\includegraphics[keepaspectratio,width=0.45\textwidth]{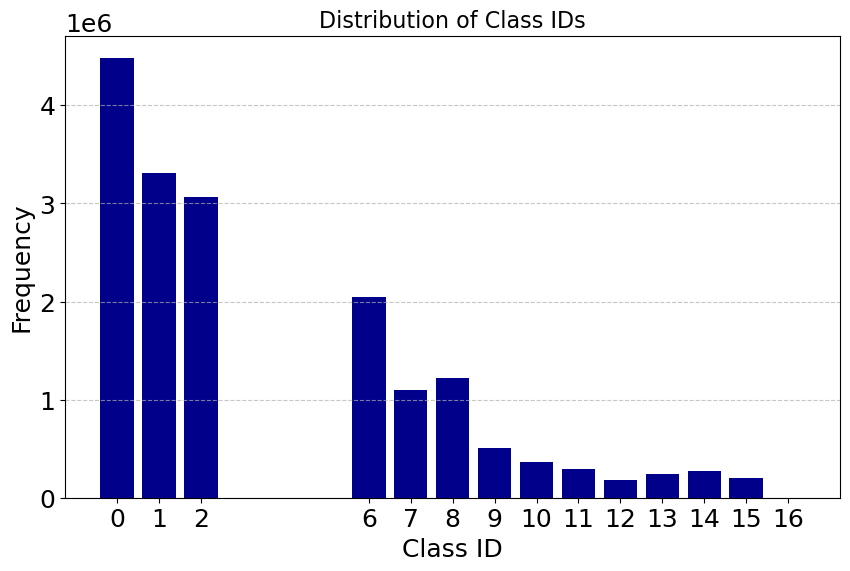}
\caption{Distribution of Class IDs. A total of sixteen objects are tracked by YOLOv8 model.}
\label{fig:distribution}
\end{center}
\end{figure}

Figure~\ref{fig:distribution} displays the distribution of the number of frames for each class ID corresponding to the sixteen objects listed in Table~\ref{tab:table_toolname} for GJ dataset. We used YOLOV8, as described in Section~\ref{sec:objectdetection}, to extract and track surgical instruments. Each instrument is segmented into three parts, each assigned a unique ID: the shaft, wrist, and jaw, except for the ``laptool''. It is important to note that in our dataset, instrument number two (``instrument2'') is identical to instrument number three (``instrument3'') because they are both needle drivers. Therefore, they are combined under the same object category. As a result, the class IDs 3, 4, and 5 are merged into IDs 2, 1, and 0.

Since the tools have multiple parts and few tools are identical to others, to avoid confusion in our analysis of tools for object detection, we define the following naming conventions:
\begin{itemize}
    \item ``One Object'': Refers to tracking part 1 (the jaws) of the Needle Driver located on the right-hand side of the screen.  
    \item ``Two Objects'': Refers to tracking part 1 (the jaws) of each of the two Needle Drivers-one located on the left-hand side and the other on the right-hand side of the screen.
    \item ``Four Objects'': Refers to tracking part 1 (the jaws) and part 2 (the wrist) of each of the two Needle Drivers-one located on the left-hand side and the other on the right-side of the screen.
    \item ``Six Objects'': Refers to tracking part 1 (the jaws), part 2 (the wrist), and part 3 (the shaft) of each of the two Needle Drivers-one located on the left-hand side and the other on the right-side of the screen.
\end{itemize}

We track three parts of the tool rather than the whole tool, which captures the detailed movement of the robotic tools that could improve the accuracy of the detection of key frames.

In addition, the image is divided into two halves by a vertical line at the center. At any given moment in the video, a tool is considered the left-hand tool if its centroid is located on the left half and the right-hand tool if its centroid is located on the right half.

Please note that, for the identification of other tools if needed, we will use a combination of the tool name and the part number (e.g., grasper part 1).
\definecolor{Silver}{rgb}{0.752,0.752,0.752}
\begin{table}[tb!]
\centering
\caption{Tool name. The name of tools along with its parts and ids.}
\label{tab:table_toolname}
\begin{tblr}{
  row{1} = {Silver},
  hlines,
  vlines,
}
ID & Instrument Part & Tool Name                 \\
0  & instrument3part1 & Needle Driver             \\
1  & instrument3part2 & Needle Driver             \\
2  & instrument3part3 & Needle Driver             \\
3  & instrument2part3 & Needle Driver             \\
4  & instrument2part2 & Needle Driver             \\
5  & instrument2part1 & Needle Driver             \\
6  & laptool          & Irrigator                 \\
7  & instrument4part1 & Forcep                    \\
8  & instrument1part1 & Grasper                   \\
9  & instrument1part2 & Grasper                   \\
10 & instrument1part3 & Grasper                   \\
11 & instrument4part2 & Forcep                    \\
12 & instrument5part3 & Monopolar Curved Scissors \\
13 & instrument5part2 & Monopolar Curved Scissors \\
14 & instrument5part1 & Monopolar Curved Scissors \\
15 & instrument4part3 & Forcep                    
\end{tblr}
\end{table}
\subsection{Hyper-parameter}
Our proposed methods were developed using a PyTorch framework and implemented on a Windows (R) 11 Enterprise system equipped with an AMD Ryzen\textsuperscript{TM} Threadripper\textsuperscript{TM} PRO 3995WX 64-cores CPU @ 2.70 GHz. We trained our model with two Nvidia RTX A6000 GPUs.  

The experiments were performed based on the configurations specified in Table~\ref{tab:table_hyperparameter}. The hyperparameters, such as image size, batch size, and clip length, were optimized to make full use of GPU memory. Please note that the threshold values for KAFR are determined based on the desired number of extracted frames. Instead of selecting a value between 0 and 1 as in the Equation~\ref{eq:f1}, we choose a number of frames by a percentage of the total number of training samples and adjust the threshold accordingly.
% \begin{table}[!tbhp]
% \centering
% \caption{\label{tab:table_hyperparameter}Hyperparameters.}
% \begin{tabular}{lc} 
% Hyperparameter                    &         Value                     \\
% \hline
% X3D CNN                 &                              \\
% Image size                        & $300\times300$                        \\
% Optimizer                         & AdamW                        \\
% Schedule                          & StepLR                       \\
% Learning rate                     & 0.001                        \\
% Decay rate                        & 0.7                          \\
% Batch size                        & 64                           \\
% Epochs                            & 100                          \\
% Early stop                        & 10                           \\
% Clip length                       & 16                           \\  
% Number frame per step             & 250                          \\ 
%                                   & \multicolumn{1}{l}{}         \\
% KAFR                              & \multicolumn{1}{l}{}         \\
% Threshold                         & Flexible                     \\
%                                   & \multicolumn{1}{l}{}         \\
% \hline
% \end{tabular}
% \end{table}

% \usepackage{caption}
% \usepackage{colortbl}

\begin{table}[!tbhp]
\centering
\caption{\label{tab:table_hyperparameter}Hyperparameters.}
\begin{tabular}{lc} 
\hline
\rowcolor[rgb]{0.753,0.753,0.753} Hyperparameter & Value                 \\ 
\hline
X3D CNN                                          &                       \\
Image size                                       & $300\times300$        \\
Optimizer                                        & AdamW                 \\
Schedule                                         & StepLR                \\
Learning rate                                    & 0.001                 \\
Decay rate                                       & 0.7                   \\
Batch size                                       & 64                    \\
Epochs                                           & 100                   \\
Early stop                                       & 10                    \\
Clip length                                      & 16                    \\
Number frame per step                            & 250                   \\
                                                 & \multicolumn{1}{l}{}  \\
KAFR                                             & \multicolumn{1}{l}{}  \\
Threshold                                        & Flexible              \\
\hline
\end{tabular}
\end{table}

For the generation of optical flow images, we implemented the Farneback estimation using the calcOpticalFlowFarneback function from OpenCV. We set the number of pyramid levels to $3$, the pyramid scale to $0.5$, the window size to $15$, the number of iterations to $3$, the pixel neighborhood size to $5$, and a standard deviation of $1.2$. The process includes converting frames to grayscale, using the calcOpticalFlowFarneback function to compute flow vectors, and saving the magnitude as an image. The estimated optical flow may be used as a secondary input to our X3D CNN model.
\subsection{Evaluation Metrics}
We adopt traditional evaluation metrics, namely accuracy and F1 score, to analyze the performance of the proposed models.
\begin{enumerate}
    \item Accuracy: it computes the percentage of correct predictions over all predictions made on the test data; that is, it represents the proportion of predictions that were correctly classified relative to all predictions, $\text{Accuracy} = (\text{Number of correct predictions}) / (\text{All predictions}).$
    \item F1 score: it establishes a uniform metric for evaluating the precision and recall by taking into account both false positives and false negatives, $\text{F1 score} = 2 \times (\text{Precision} \times \text{Recall}) / (\text{Precision} + \text{Recall}).$
    \item Accuracy change (\%):  in our experiments, we use relative change instead of absolute change when referring to the increase or decrease in the accuracy of our methods. The percentage relative change is defined as follows: $\text{Accuracy change (\%)} = (\text{New Accuracy} - \text{Original Accuracy}) / (\text{Original Accuracy}) \times 100$.
    \item F1 score change (\%): the score is computed in the same manner as the accuracy change.
\end{enumerate}
\section{Classification Results}
\label{sec:result}
\subsection{Adaptive 1}
\begin{figure*} [hbtp!]
\begin{center}
\includegraphics[keepaspectratio,width=0.95\textwidth]{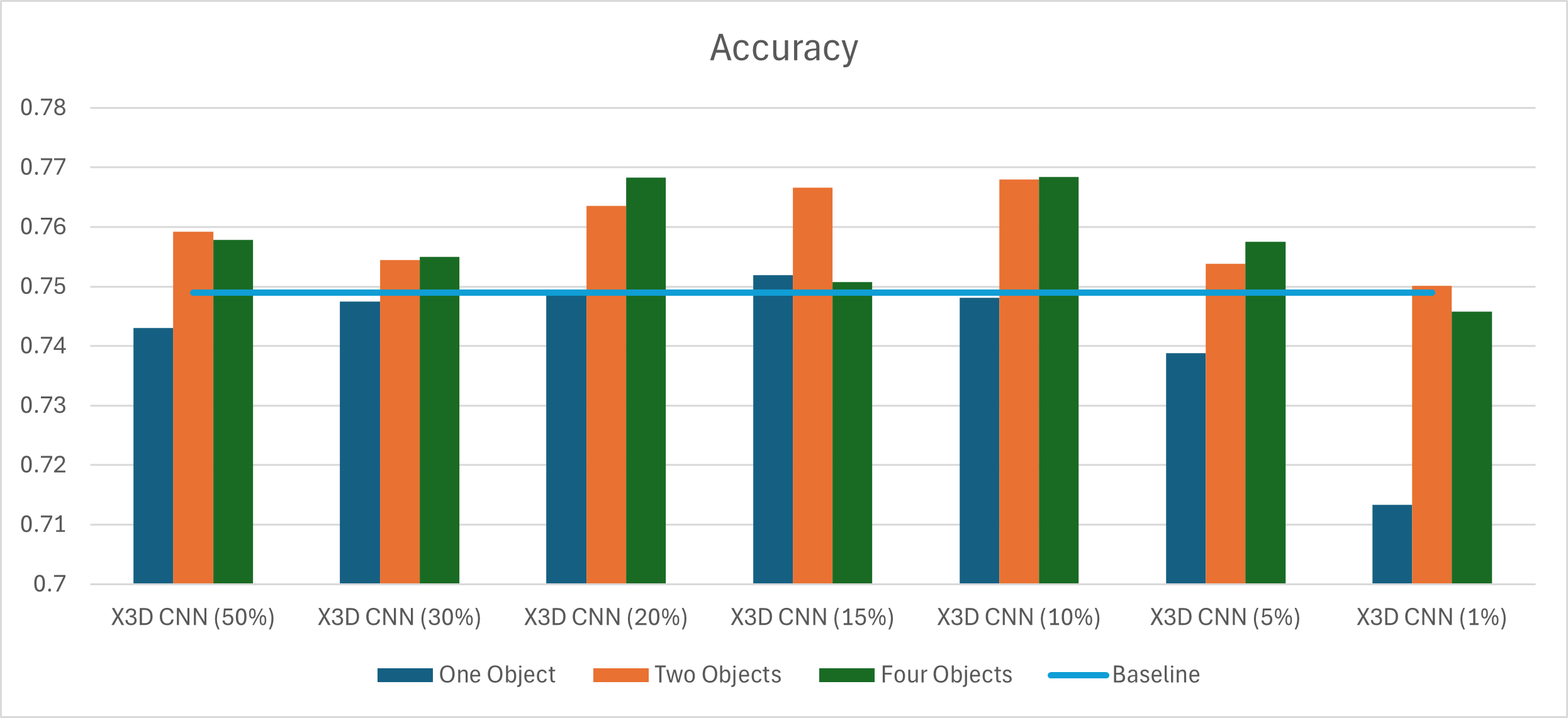}
\caption{Performance Using Velocity.}
\label{fig:tools}
\end{center}
\end{figure*}

In this experiment, we initially focused on tracking velocity (displacement over time) using one object—specifically, the right-hand tool, as most procedures were performed by right-hand dominant surgeons. However, this method frequently fell short of surpassing the baseline accuracy of $0.749$. This ``baseline'' represents the result obtained using the research settings described in prior work (Section~\ref{sec:extension}). The turning point occurred when we allocated $15\%$ of our data, resulting in an improvement that exceeded the baseline by $0.38\%$ (Figure~\ref{fig:tools} and Table~\ref{tab:table_velocity}).

Building upon these promising findings, we expanded our experiments to incorporate velocity tracking from two objects. The result was remarkable; with just a $10\%$ data allocation, we achieved a $2.53\%$ increase in accuracy. Our observations indicated that the method was most effective with sufficient data reduction ($20\%$, $15\%$, and $10\%$ of total frames), suggesting that critical frames were being effectively captured. However, significant reductions in data ($5\%$ and $1\%$ of total frames) led to decreased accuracy, likely due to insufficient data for robust model training.

Subsequent studies involving additional objects yielded slight accuracy gains ($30\%$, $20\%$, $10\%$, and $5\%$) and, in some cases, resulted in reduced accuracy ($50\%$, $15\%$, and $1\%$ data allocations). Nevertheless, we attained the best accuracy of $0.7684$ by utilizing four objects with a $10\%$ frame allocation. Therefore, we chose to limit our analysis to four objects, as incorporating additional ones did not yield substantial improvements in model accuracy, despite being available in the dataset.

These outcomes indicate the important role of adaptively selecting key frames in optimizing model performance. Our Adaptive 1 method, which selects frames based on recognition of significant tool movements, not only improves accuracy but also notably reduces the dataset size. In the next section, we will compare our approach with a well-known method to highlight its competitiveness.
\definecolor{Silver}{rgb}{0.752,0.752,0.752}
\begin{table}[!tbp]
\centering
\caption{Effect of Velocity.}
\label{tab:table_velocity}
\begin{tblr}{
  row{1} = {Silver},
  row{2} = {Silver},
  column{3} = {c},
  column{4} = {c},
  cell{1}{1} = {c=2}{},
  cell{1}{3} = {c=2}{},
  cell{3}{1} = {r=7}{},
  cell{10}{1} = {r=7}{},
  cell{17}{1} = {r=7}{},
  vlines,
  hline{1-3,10,17,24} = {-}{},
  hline{4-9,11-16,18-23} = {2-4}{},
}
                                     &                & Performance     &        \\
Data                                 & Model          & Accuracy        & F1 Score    \\
{One Object}           & X3D CNN ($1\%$)  & 0.7133          & 0.6071 \\
                                     & X3D CNN ($5\%$)  & 0.7388          & 0.6806 \\
                                     & X3D CNN ($10\%$) & 0.7481          & 0.6857 \\
                                     & X3D CNN \textbf{($15\%$)} & \textbf{0.7519}          & 0.6830 \\
                                     & X3D CNN ($20\%$) & 0.7495          & 0.6829 \\
                                     & X3D CNN ($30\%$) & 0.7475          & 0.6733 \\
                                     & X3D CNN ($50\%$) & 0.7430          & 0.6577 \\
{Two Objects}  & X3D CNN ($1\%$)  & 0.7501          & 0.6044 \\
                                     & X3D CNN ($5\%$)  & 0.7538          & 0.6529 \\
                                     & X3D CNN \textbf{($10\%$)} & \textbf{0.7680}          & 0.7129 \\
                                     & X3D CNN ($15\%$) & 0.7535          & 0.6781 \\
                                     & X3D CNN ($20\%$) & 0.7635          & 0.6812 \\
                                     & X3D CNN ($30\%$) & 0.7544          & 0.6810 \\
                                     & X3D CNN ($50\%$) & 0.7592          & 0.6769 \\
{Four Objects} & X3D CNN ($1\%$)  & 0.7458          & 0.6002 \\
                                     & X3D CNN ($5\%$)  & 0.7575          & 0.6816 \\
                                     & X3D CNN \textbf{($10\%$)} & \textbf{0.7684} & 0.6900 \\
                                     & X3D CNN ($15\%$) & 0.7666          & 0.6943 \\
                                     & X3D CNN ($20\%$) & 0.7683          & 0.6906 \\
                                     & X3D CNN ($30\%$) & 0.7550          & 0.6906 \\
                                     & X3D CNN ($50\%$) & 0.7578          & 0.6754 
\end{tblr}
\end{table}
\subsection{Comparison}
\label{sec:comparison}
This experiment offers a comparative analysis between our proposed method, KAFR, and the widely used MSE metric. MSE, commonly employed in tasks such as detection of moving objects and loss function measurement, evaluates similarity using all pixel-wise information within an image. In contrast, KAFR focuses only on specific pixels (the centroids of the tracked tools), disregarding other pixel data. This distinction highlights a key difference between the two methods: MSE provides a holistic pixel-based analysis, while KAFR concentrates exclusively on the identified tools.

\begin{figure} [hbtp!]
\begin{center}
\includegraphics[keepaspectratio,width=0.45\textwidth]{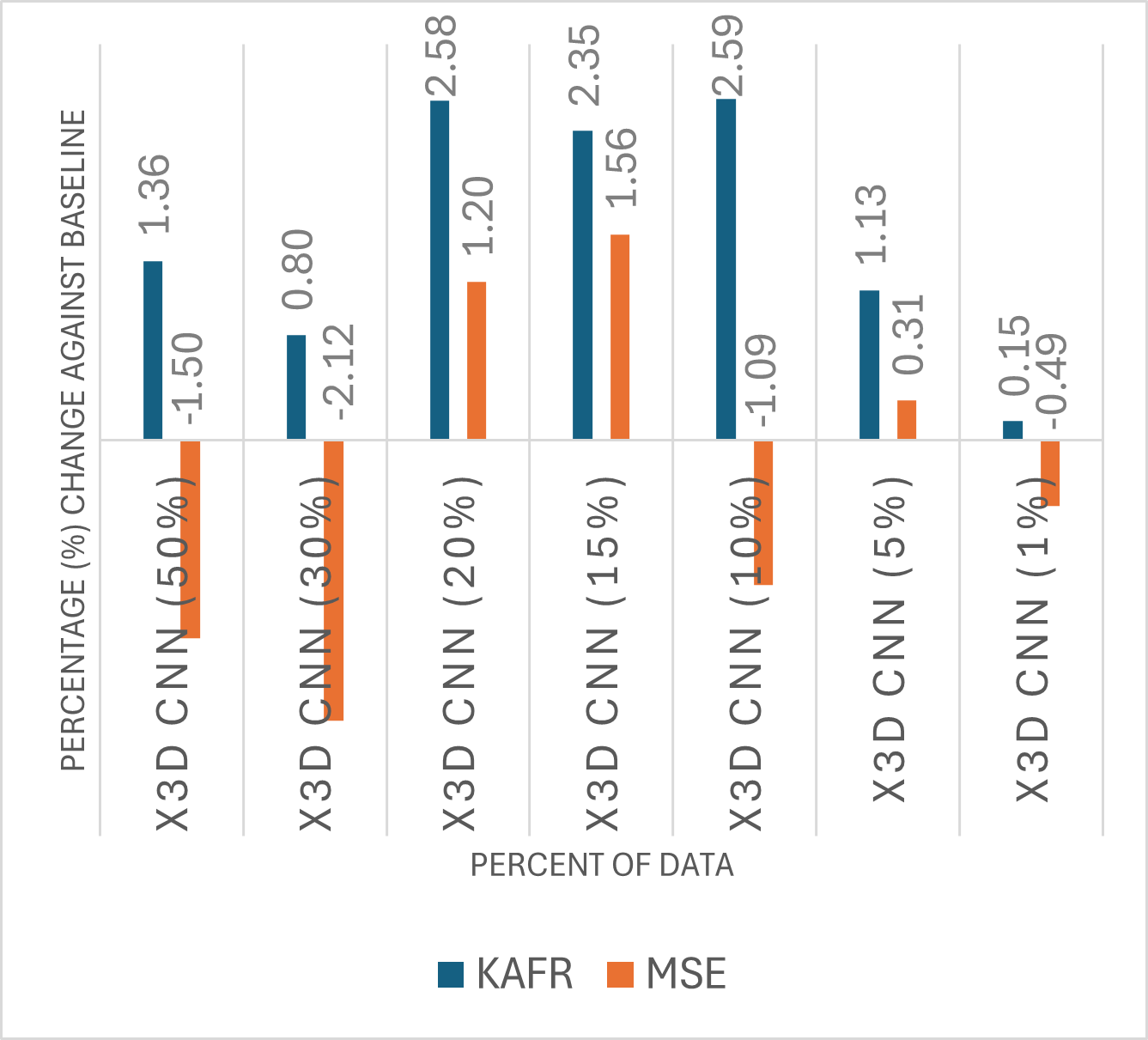}
\caption{KAFR vs. MSE.}
\label{fig:mse}
\end{center}
\end{figure}

Figure~\ref{fig:mse} illustrates the performance of KAFR and MSE relative to a baseline, indicating whether their accuracy has improved or decreased. KAFR uses the best results obtained from combinations of different objects, as discussed in the previous section. For example, with X3D CNN (10\%), we selected the best result from two objects, which was $0.7680$, as it outperformed the results from one and four objects (Table~\ref{tab:table_velocity}) and computed the accuracy change accordingly. In contrast, MSE uses the same portions of the dataset but processes all pixel data. As we can see, MSE occasionally outperforms the baseline with $20\%$, $15\%$, and $5\%$ data allocations but exhibits notable limitations with $50\%$, $30\%$, $10\%$, and $1\%$ data allocations, implying vulnerability to background noise interference that impacts its performance. It should be emphasized that in surgical videos, the background, which typically consists of deformable organs and connective tissues, is constantly moving, making it harder to extract useful information from tool movements. On the other hand, KAFR outperforms the baseline in most cases, achieving improvements of up to approximately $2.59\%$ (X3D CNN ($10\%$)). This demonstrates its effectiveness in selectively tracking and utilizing tool data to identify key frames. In addition to comparing to the baseline, when both methods are put together, KAFR outperforms the MSE.
\subsection{Adaptive 2}
\begin{figure} [tbp!]
\begin{center}
\includegraphics[keepaspectratio,width=0.45\textwidth]{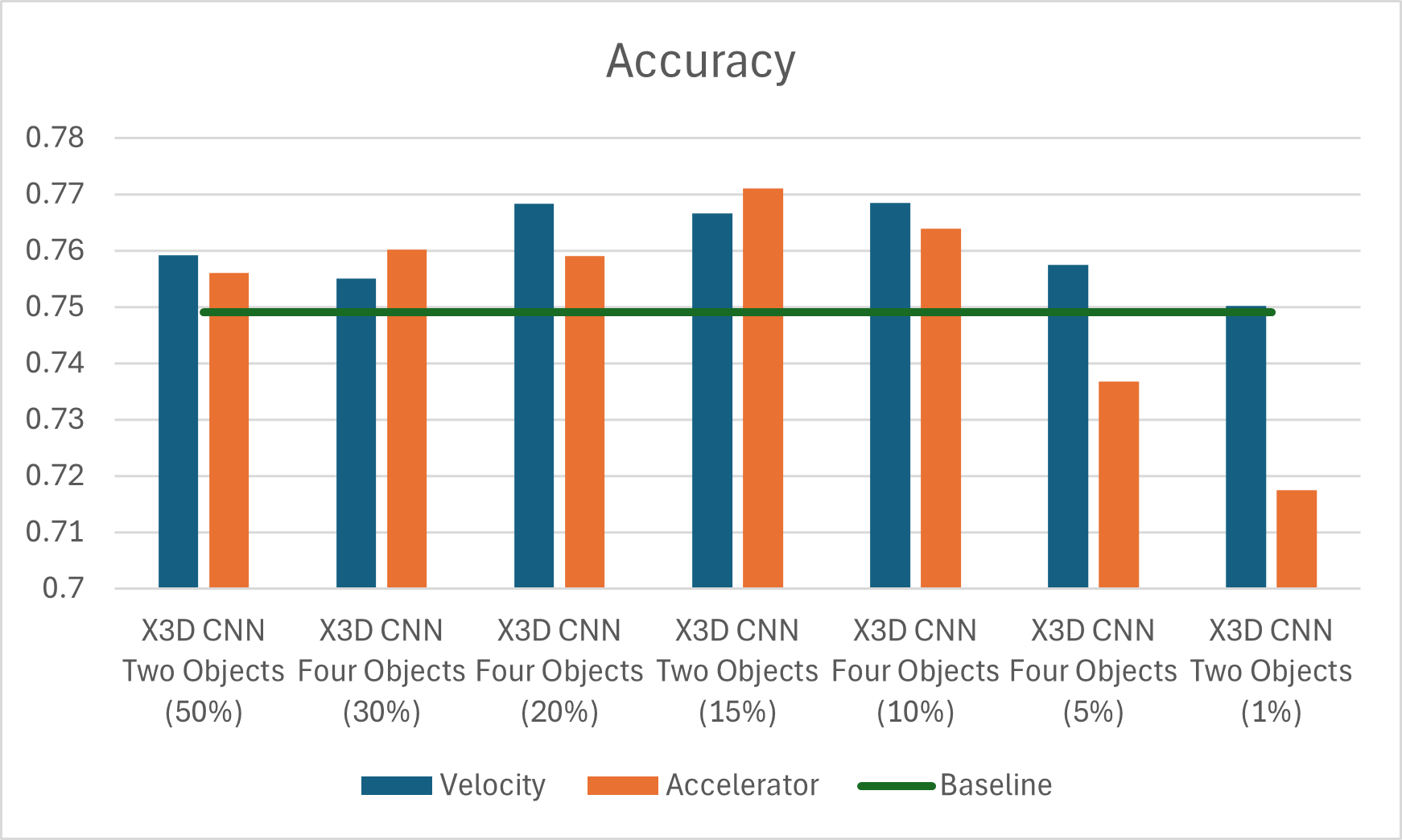}
\caption{Comparison of Velocity, Acceleration, and Baseline.}
\label{fig:accelerator}
\end{center}
\end{figure}

An alternative method to detect tool movement is to analyze the variations of the rate of change of velocity over time, referred to as acceleration. These accelerations may be a sign of significant surgical maneuvers made by surgeons. Looking at Figure~\ref{fig:accelerator}, a trend emerges similar to that of the velocity (Figure~\ref{fig:tools}), where, in most cases, the accuracy outperforms the baseline. However, the accuracy decreased when the number of frames was reduced to $5\%$ and $1\%$. Although the significance of accelerations in detecting crucial frames may not match that of velocity, we achieved better accuracy ($0.771$) using a $15\%$ data allocation for accelerations.
\subsection{Two-stream Convolutional Networks}
\begin{figure*} [tbhp!]
\begin{center}
\includegraphics[keepaspectratio,width=0.95\textwidth]{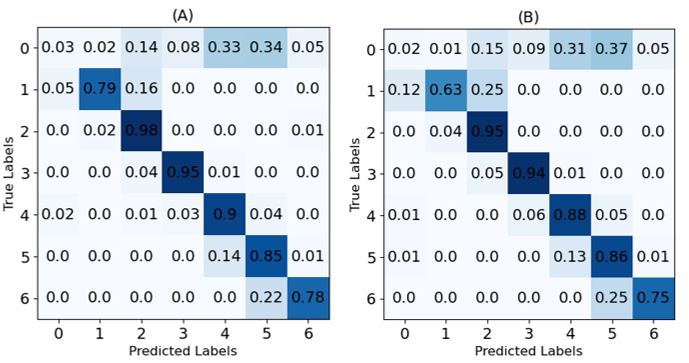}
\caption{Confusion Matrix. (A) X3D CNN Two Objects Acceleration (15\%) and (B) X3D CNN Four Objects
Velocity (10\%). The numbers indicate: ’1.1 Stay suture’ - ’1’, ’1.2 Inner
running suture’ - ’2’, ’1.3 Enterotomy’ - ’3’, ’2.2 Inner running suture’ - ’4’, ’3.1 Inner Layer of Connell’ - ’5’, ’4.1 Outer Layer of Connell’ -
’6’, ’Idle time’ - ’0’, respectively.}
\label{fig:cm}
\end{center}
\end{figure*}

Table~\ref{tab:table_ensemble} shows the results achieved using the two best candidates from the previous section, namely Acceleration with Two Objects using $15\%$ of the data and Velocity with Four Objects using $10\%$ of the data. The outputs of the two channels (Optical Flow and RGB) are ensembled. As we can see, the accuracy is further improved to $0.7814$ ($4.32\%$) and slightly increased for the F1 score ($0.16\%$). In addition, the confusion matrices are plotted in Figure~\ref{fig:cm}.
\definecolor{Silver}{rgb}{0.752,0.752,0.752}
\begin{table}[!tbh]
\centering
\caption{Two-stream Convolutional Networks.}
\label{tab:table_ensemble}
\begin{tblr}{
  row{1} = {Silver},
  row{2} = {Silver},
  column{3} = {c},
  column{5} = {c},
  cell{1}{1} = {r=2}{},
  cell{1}{2} = {c=2}{c},
  cell{1}{4} = {c=2}{c},
  cell{3}{2} = {c},
  cell{3}{4} = {c},
  cell{4}{2} = {c},
  cell{4}{4} = {c},
  cell{5}{2} = {c},
  cell{5}{4} = {c},
  vlines,
  hline{1,3-6} = {-}{},
  hline{2} = {2-5}{},
}
Type         & {X3D CNN Two Objects \\Acceleration ($15\%$)} &                 & {X3D CNN Four Objects \\Velocity ($10\%$)} &          \\
             & Accuracy                                      & F1 Score        & Accuracy                                   & F1 Score \\
Optical Flow & 0.7291                                        & 0.6496          & 0.7175                                     & 0.6304   \\
RGB          & 0.7438                                        & 0.6747          & 0.7532                                     & 0.6750   \\
Ensemble     & \textbf{0.7814}                               & 0.7141 & 0.7666                                     & 0.6794   
\end{tblr}
\end{table}
\begin{figure*} [tbhp!]
\begin{center}
\includegraphics[keepaspectratio,width=0.95\textwidth]{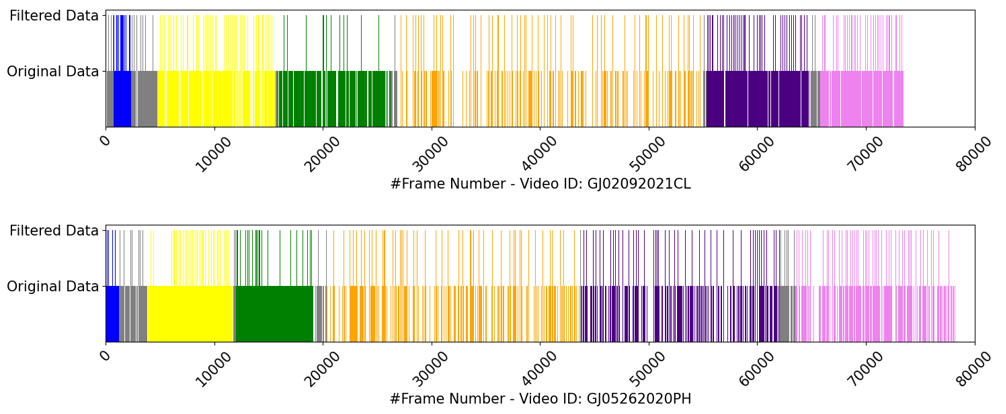}
\caption{Phase Segmentation. Two distinct videos are sampled to show KAFR performance. The color codes are: '1.1 Stay suture' - 'Blue', '1.2 Inner running suture' - 'Yellow', '1.3 Enterotomy' - 'Green', '2.2 Inner running suture' - 'Orange', '3.1 Inner Layer of Connell' - 'Indigo', '4.1 Outer Layer of Connell' - 'Violet', 'Idle time' - 'Gray', respectively. Filtered Data refers to data composed of key frames.}
\label{fig:active}
\end{center}
\end{figure*}

Figure~\ref{fig:active} illustrates the results (X3D CNN Two Objects Acceleration ($15\%$)) using two distinct videos. As shown in the figure, six phases are segmented along with idle time (in gray). The frame indices for the original data (showing gaps between frames resulting from undersampling and oversampling techniques, as more samples are extracted for shorter phases and fewer samples for longer phases) and the filtered indices after using KAFR are displayed. In addition, Table~\ref{tab:table_phasereducerate} provides details on the total number of frames and the percentage of frames filtered for these samples. The table also shows the number of duplicated frames resulting from upsampling.

\definecolor{Silver}{rgb}{0.752,0.752,0.752}
\begin{table}[tbt!]
\centering
\caption{Total frames and percentage of frames after filtering. The phases are: Phase 1 - '1.1 Stay suture', Phase 2 - '1.2 Inner running suture', Phase 3 - '1.3 Enterotomy', Phase 4 - '2.2 Inner running suture', Phase 5 - '3.1 Inner Layer of Connell', Phase 6 - '4.1 Outer Layer of Connell', Phase 0 - 'Idle time', respectively. 'Dup' is the abbreviation for duplication (number of repeated frames).}
\label{tab:table_phasereducerate}
\begin{tblr}{
  cells = {c},
  row{1} = {Silver},
  row{2} = {Silver},
  row{3} = {Silver},
  cell{1}{1} = {r=3}{},
  cell{1}{2} = {c=4}{},
  cell{2}{2} = {c=2}{},
  cell{2}{4} = {c=2}{},
  vlines,
  hline{1,4-12} = {-}{},
  hline{2-3} = {2-5}{},
}
Phase & Video ID     &                   &                       &                   \\
               & GJ02092021CL &                   & GJ05262020PH &                   \\
               & Total Frames & Filtered & Total Frames & Filtered \\
Phase 0        & 248                   & $11.69\%$           & 248                   & $15.32\%$           \\
Phase 1        & 250                   & $9.20\%$            & 250                   & $4.00\%$            \\
Phase 2        & 250                   & $19.20\%$           & 250                   & 1$4.80\%$           \\
Phase 3        & 250                   & $7.60\% $           & 250                   & $12.00\%$           \\
Phase 4        & 250                   & $24.40\%$           & 250                   & $18.80\%$           \\
Phase 5        & 250                   & $17.20\%$           & 250                   & $17.20\%$           \\
Phase 6        & 227                   & $16.30\%$           & 250                   & $21.20\%$  \\
Dup            & 138                   &                   & 144                   &          \\  
\end{tblr}

\end{table}
The KAFR tends to skip frames at the beginning and end of a phase, e.g., the first frames in gray color and the end frames in pink color in the upper plot of Figure~\ref{fig:active}. It is also observed that in the Enterotomy phase of the upper figure, the gap for filtered data appears longer. This occurs because, in this specific instance, the surgeon performed the cutting on the right hand, which operated the tool that was not tracked, while the left hand, which was tracked, moved more slowly or, at times, disappeared from the screen. As a result, fewer key frames were detected compared to other phases.
\section{Extension}
\label{sec:extension}
In this section, we apply our proposed methods, Adaptive 1 and Adaptive 2, to the PJ dataset (Section~\ref{sec:pj_dataset}). For consistency, we maintained the same model architecture (X3D CNN) and optimized settings as detailed in previous research~\cite{al2024development}. With this setup, the best model achieved an accuracy of $0.8801$ and an F1 score of $0.8534$, which we use as the baseline for comparison with our findings.

\begin{figure} [tbhp!]
\begin{center}
\includegraphics[keepaspectratio,width=0.45\textwidth]{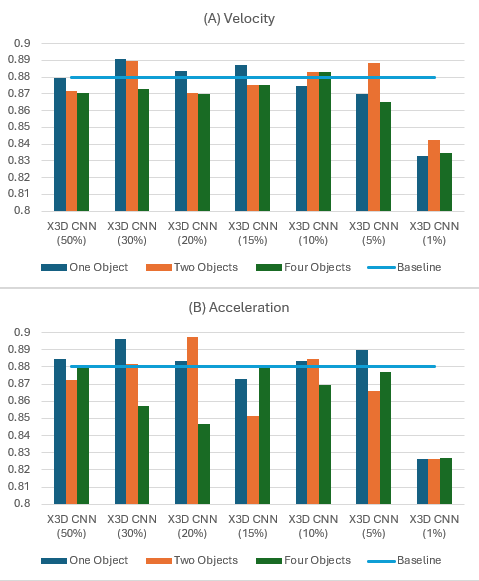}
\caption{Accuracy on PJ Dataset. (A) Adaptive 1 -- Velocity. (B) Adaptive 2 -- Acceleration.} 
\label{fig:pj_dataset}
\end{center}
\end{figure}

Figure~\ref{fig:pj_dataset} presents a comparison of the results obtained by our proposed methods against the baseline. In this evaluation, we examined the effect of varying the training set size, using increments of $1\%$, $5\%$, $10\%$, $15\%$, $20\%$, $30\%$, and $50\%$ of the total training data. Our study also explored the impact of tracking data from multiple objects on model accuracy. Specifically, we analyzed scenarios where tracking was applied using one, two, and four objects in the same manner as with the GJ dataset.

Our best results outperformed the baseline, demonstrating the effectiveness of our adaptive methods. The highest accuracy for Adaptive 1 reached $0.8909$ (a $1.22\%$ increase) when using velocity data from one object with $30\%$ of the data. For Adaptive 2, the top accuracy was $0.8976$ (a $1.98\%$ increase), achieved with acceleration data from two objects and $20\%$ of the data. The accuracy is increased to $0.8982$ ($2.05\%$) and the F1 score to $0.8751$ ($2.54\%$) with the ensembling method, employing Adaptive 2 with two objects approach on $20\%$ of the data.
\section{State-of-the-Art Methods}
\label{sec:stateoftheart}
% Previously, our group conducted research on the PJ dataset using the X3D CNN model. In this study, we build upon that work by introducing a new method, Kinematics Adaptive Frame Recognition, to selectively capture key frames rather than using all frames. This approach reduces the number of frames by fivefold and improves accuracy by approximately 2\%.
%
\begin{table}[bth!]
\centering
\caption{State-Of-The-Art Methods on Distinct Datasets.}
\begin{tabular}{|l|l|c|}
\hline
\rowcolor[rgb]{0.753,0.753,0.753}\textbf{Dataset Name} & \textbf{Method}       & \textbf{Accuracy ($\%$)} \\ \hline
Cholec80              & TeCNO~\cite{czempiel2020tecno}                 & 88.56             \\ \cline{2-3}
                      & Endonet + HHMM~\cite{twinanda2016endonet}           & 81.70             \\ \cline{2-3}
                      & LoViT~\cite{liu2025lovit}                 & 92.40             \\ \cline{2-3}
                      & ST-ERFNet~\cite{jin2020multi}             & 86.07             \\ \cline{2-3}
                      & MTRCNet-CL~\cite{jin2020multi}        & 87.40             \\ \hline
Cholec51              & TeCNO~\cite{czempiel2020tecno}                 & 87.34             \\ \hline
CATARACTS             & SurgPLAN~\cite{luo2024surgplan}              & 83.10             \\ \cline{2-3}
                      & M2F + GCN~\cite{holm2023dynamic}             & 76.86             \\ \hline
M2cai16               & LAST~\cite{tao2023last}                  & 91.50             \\ \hline
PJ                    & X3D CNN~\cite{al2024development}               & 88.01             \\ \hline \hline
PJ                    & KAFR                  & 89.82                  \\ \hline
GJ                    & KAFR                   & 78.14                  \\ \hline
\end{tabular}
\label{tab:table_TSOTA}
\end{table}

As seen in Table~\ref{tab:table_TSOTA}, the TeCNO represents a Multi-Stage Temporal Convolutional Network (MS-TCN) that uses a hierarchical prediction refinement approach for surgical phase recognition. It employs causal, dilated convolutions to provide a large receptive field, enabling smooth, online predictions even during ambiguous transitions in surgical phases. In contrast to standard causal MS-TCNs, the TeCNO restricts each prediction to rely only on current and past frames, enhancing real-time application by reducing dependence on future frames. Applying this method to the Cholec80 and Cholec51 datasets yielded accuracies of $88.56\%$ and $87.34\%$, respectively. Although TeCNO arguably surpasses many state-of-the-art LSTM approaches, training ResNet50 frame-by-frame without temporal context may reduce effectiveness by losing crucial frame-to-frame relationships.

The Long Video Transformer (LoViT) takes another approach by focusing on the extraction of temporal spatial features and modeling phase transitions. It utilizes a multiscale temporal aggregator that integrates L-Trans modules based on self-attention to capture short-term, fine-grained details and a G-Informer module utilizing ProbSparse self-attention to examine long-term temporal dynamics. These features are fused through a multiscale temporal head, which employs phase transition-aware supervision for accurate classification. The model achieves an accuracy of $92.40\%$ on the Cholec80 dataset. However, it does not take into consideration the dynamics of surgical tool movement, such as velocity and acceleration, which are addressed in Adaptive 1 and Adaptive 2. Additionally, the source code was unavailable at the time of writing this manuscript, preventing a fair comparison by applying the LoViT model to our datasets.

The previous research conducted on the PJ dataset using the X3D CNN model employed the Uniform Frame Sampling method (6 fps)~\cite{al2024development}, which is neither adaptive nor explainable. In this study, we used the provided source code and \textbf{improved}  upon that work by introducing a new method, Kinematics Adaptive Frame Recognition, to selectively capture key frames rather than using all frames. This approach reduces the number of frames by fivefold and improves accuracy and F1 score by approximately $2\%$. In single-task operating systems, this reduction proportionally decreases the time required for training the model. Theoretically, in real multitasking operating systems, where multiple tasks can run simultaneously, the time savings align closely with the reduction in data size. However, in near-multitasking operating systems, such as Windows (R) operating systems, the time reduction may not scale linearly because the main processes are influenced by competing user and system processes, leading to potential bottlenecks. In our experiments, the runtime required for one epoch of the total training data was approximately 19'15'', while for $10\%$ of the data, it was roughly 5'24''. Although the computation time was reduced, it did not scale directly with the reduction in data size. While an in-depth analysis of the effectiveness of multitasking computing systems falls beyond the scope of this research, it is important to note that reducing the number of frames would not only significantly decrease the storage requirements but also improve computational efficiency, making the method more resource-efficient and practical for large-scale applications.

In terms of key frames recognition, our proposed method differs from a recent paper~\cite{loukas2018keyframe}, although both aim to detect key frames within a video. First, our approach uses YOLOv8 to detect the presence of these tools in the scene and track their locations across frames. In contrast, the other approach relies on color, motion, and texture and uses a less fine-grained method for motion detection, tracking all moving objects without focusing on specific tools. Second, our goal is to use key frames for phase segmentation, which requires more frames than the other approach, whose objective is merely to select representative frames for summarizing the video, requiring fewer frames. Additionally, we use the X3D CNN model to analyze the temporal information of 2-D frame sequences instead of the hidden Markov model.

Last but not least, traditional compression techniques, such as those used by the Moving Picture Experts Group (MPEG) standard~\cite{le1991mpeg}, primarily focus on reducing file size by leveraging spatial and temporal redundancies in video data without considering the semantic relevance of specific content. While effective for general-purpose data compression, these methods fail to account for the nuanced actions performed by surgeons, which are critical for surgical analysis. In contrast, our proposed method not only reduces the dataset size but also preserves the semantic significance of surgical actions by tracking the movements of tools. By incorporating this context-aware approach, it achieves improved performance alongside data size reduction. This dual benefit highlights the advantage of the proposed method for specialized applications like surgical skill assessment.
\section{Discussion}
\label{sec:discussions}
In surgery, operative videos are often hierarchically segmented into phases, steps, and tasks to describe the procedure at various levels of detail~\cite{meireles2021sages}. Each segment is typically defined by a start time and an end time and often spans a considerable duration (from minutes to hours). Often, many portions of the procedure may be idle, for instance, when tools are exchanged or when the camera is cleared of condensation. Moreover, surgical tools are critical to identify in each phase as they are used for performing the surgery, including for retraction, resection, blunt and sharp dissection, cauterizing bleeding, and irrigating the surgical field. 
In this work, tracking the surgical tools and using KAFR to adaptively select key frames for phase detection resulted in improvements that exceeded our expectations. Initially, we anticipated that using KAFR would involve a trade-off between dataset size and performance. In other words, we expected the accuracy to decrease to some extent for the reduction of data size~\cite{nguyen2019advanced,phong2019improvement}. However, the results indicate otherwise: accuracy can improve even as the dataset size decreases. As demonstrated in the previous section, when the left-hand tool moved slowly, fewer frames were selected. The removed frames had little effect on recognizing the actions~\cite{phong2018action,nguyen2023video} performed by the surgeon. Since the model was trained using a frame-by-frame approach, the removed frames introduced confusion during training. Eliminating these frames ultimately resulted in the improvement.

One potential issue with our approach is that using YOLO to track surgical tools often results in loss of tracking, either due to tools being retracted or occluded. When the tool is visible or tracked again, a new tracking number is automatically assigned to it, as if it were an instance of a new object, needing to continuously monitor the tracking id and assign them to the appropriate tools during post-processing. In our experiment, left-hand and right-hand tools are detected based on their centroids relative to a vertical line at the center of the screen. If we track only one tool on either side of the screen, it may move from one side to the other, which can lead to loss of tracking. However, since these tools tend to stay near the center of the screen--convenient for surgeons to focus on--the right-hand or left-hand tool typically remains on its respective side, minimizing the effects. Additionally, because we track tools on both the left and right sides of the screen, the system continues tracking even when a tool crosses to the opposite side. To mitigate this issue, identical tools could be painted with different colors or labeled with unique numbers. However, as our videos were collected for a retrospective study, we were unable to implement or evaluate this setting. This limitation highlights an area for potential future improvement.

Another limitation is that our method assumes the endoscopic camera remains stationary. Changes in the camera view (zoom in and out or excessive movements) can affect the computation of distances, reducing the accuracy of the method. Frequent camera movement similarly impacts other AI models, making this a broader challenge in the field. Our datasets were obtained from robotic procedures where the camera movements are limited compared to laparoscopic procedures where it is controlled manually.

Finally, our methods (Adaptive 1 and Adaptive 2) assume the presence of surgical tools. While we tested it on two datasets, GJ and PJ, and it can be applied to a broader range of datasets, the approach remains restricted to surgical fields. In other domains, the method may face challenges if specific objects are unavailable for tracking.
\section{Conclusion}
\label{sec:conclusion}
In conclusion, the growing interest in utilizing AI to automate surgical phase segmentation analysis has prompted extensive research efforts. Surgical videos, essential for performance assessment and analysis, pose challenges to AI models due to their length. However, with the expected increase in video volume, there is an urgent need for innovative techniques to address these challenges effectively.

In this study, we introduced Kinematics Adaptive Frame Recognition, a novel method aimed at reducing dataset size and computation time by removing redundant frames while preserving critical information to improve accuracy. The KAFR approach consists of three primary phases: i) Use a YOLOv8 model to detect surgical tools in the scene; ii) Compute similarities between consecutive frames by analyzing spatial positions and velocities of detected tools; iii) Train an X3D CNN to perform phase segmentation based on extracted frames.

We evaluated the efficacy of KAFR on both the GJ and PJ datasets over a span of 10–12 weeks. By adaptively selecting relevant frames, we achieved a tenfold reduction in the number of frames while improving accuracy by more than $4\%$ and the F1 score by $0.16\%$ on the GJ dataset and a fivefold reduction in frames with approximately $2\%$ improvement in both accuracy and F1 score on the PJ dataset. Furthermore, we conducted a comparative analysis with the state-of-the-art method MSE to demonstrate the competitive performance and efficiency of our approach. Moreover, KAFR can be used to enhance existing methods, providing a valuable improvement to the performance of other AI models. In addition to boosting performance, extracting key frames from a video enhances interpretability and explainability. This process simplifies complex information, clarifies the rationale behind decisions, and links outputs to specific moments. Overall, it makes video-based systems more user-friendly and trustworthy.
\section*{Acknowledgments}
The authors would like to thank Professor Bernardete Ribeiro and Dr. Francisco Antunes from the Department of Informatics Engineering, University of Coimbra, for their valuable assistance and insightful discussions. 

We acknowledge the funding from NIH/NIBIB R01: EB025247 that supported Drs. Nguyen, Garces Palacios and Sankaranarayanan.
\section*{Ethical approval} 
The institutional review board -- the ethics committee of the UT Southwestern Medical Center approved the study design and the use of GJ videos. All methods were performed in accordance with the relevant guidelines and regulations.
\section*{Data Availability}
The data that support the findings of this study may be provided upon reasonable request.
\bibliographystyle{unsrt}
\bibliography{kafr}

\begin{thebibliography}{10}

\bibitem{phong2022pso}
Huu~Phong Nguyen, Augusto Santos, and Bernardete Ribeiro.
\newblock Pso-convolutional neural networks with heterogeneous learning rate.
\newblock {\em IEEE Access}, 10:89970--89988, 2022.

\bibitem{nguyen2023video}
Huu~Phong Nguyen and Bernardete Ribeiro.
\newblock Video action recognition collaborative learning with dynamics via pso-convnet transformer.
\newblock {\em Scientific Reports}, 13(1):14624, 2023.

\bibitem{al2024development}
Amr~I Al~Abbas, Babak Namazi, Imad Radi, Rodrigo Alterio, Andres~A Abreu, Benjamin Rail, Patricio~M Polanco, Herbert~J Zeh~III, Melissa~E Hogg, Amer~H Zureikat, et~al.
\newblock The development of a deep learning model for automated segmentation of the robotic pancreaticojejunostomy.
\newblock {\em Surgical Endoscopy}, pages 1--9, 2024.

\bibitem{liu2023yolo}
Zhiguo Liu, Yuan Gao, Qianqian Du, Meng Chen, and Wenqiang Lv.
\newblock Yolo-extract: Improved yolov5 for aircraft object detection in remote sensing images.
\newblock {\em IEEE Access}, 11:1742--1751, 2023.

\bibitem{liang2015recurrent}
Ming Liang and Xiaolin Hu.
\newblock Recurrent convolutional neural network for object recognition.
\newblock In {\em Proceedings of the IEEE conference on computer vision and pattern recognition}, pages 3367--3375, 2015.

\bibitem{felzenszwalb2009object}
Pedro~F Felzenszwalb, Ross~B Girshick, David McAllester, and Deva Ramanan.
\newblock Object detection with discriminatively trained part-based models.
\newblock {\em IEEE transactions on pattern analysis and machine intelligence}, 32(9):1627--1645, 2009.

\bibitem{sun2024k}
Weirong Sun, Yujun Ma, and Ruili Wang.
\newblock k-nn attention-based video vision transformer for action recognition.
\newblock {\em Neurocomputing}, 574:127256, 2024.

\bibitem{phong2023pattern}
Huu~Phong Nguyen.
\newblock {\em Pattern Recognition: Contributions and Applications to Image Classification and Video Recognition}.
\newblock PhD thesis, Universidade de Coimbra, 2023.

\bibitem{garg2024human}
Shruti Garg, Sudhir Sharma, Sumit Dhariwal, W~Deva Priya, Mangal Singh, and S~Ramesh.
\newblock Human crowd behaviour analysis based on video segmentation and classification using expectation--maximization with deep learning architectures.
\newblock {\em Multimedia Tools and Applications}, pages 1--23, 2024.

\bibitem{grammatikopoulou2024spatio}
Maria Grammatikopoulou, Ricardo Sanchez-Matilla, Felix Bragman, David Owen, Lucy Culshaw, Karen Kerr, Danail Stoyanov, and Imanol Luengo.
\newblock A spatio-temporal network for video semantic segmentation in surgical videos.
\newblock {\em International Journal of Computer Assisted Radiology and Surgery}, 19(2):375--382, 2024.

\bibitem{li2020abnormal}
Ang Li, Zhenjiang Miao, Yigang Cen, Xiao-Ping Zhang, Linna Zhang, and Shiming Chen.
\newblock Abnormal event detection in surveillance videos based on low-rank and compact coefficient dictionary learning.
\newblock {\em Pattern Recognition}, 108:107355, 2020.

\bibitem{gao2024human}
Zhenhai Gao, Tong Yu, Fei Gao, Rui Zhao, and Tianjun Sun.
\newblock Human-like mechanism deep learning model for longitudinal motion control of autonomous vehicles.
\newblock {\em Engineering Applications of Artificial Intelligence}, 133:108060, 2024.

\bibitem{IJMESD400}
Balaram~Yadav Kasula.
\newblock {AI} applications in healthcare a comprehensive review of advancements and challenges.
\newblock {\em International Journal of Managment Education for Sustainable Development}, 6(6), 2023.

\bibitem{josiah2024artificial}
Josiah~G. Aklilu, Min~Woo Sun, Shelly Goel, Sebastiano Bartoletti, Anita Rau, Griffin Olsen, Kay~S. Hung, Sophie~L. Mintz, Vicki Luong, Arnold Milstein, Mark~J. Ott, Robert Tibshirani, Jeffrey~K. Jopling, Eric~C. Sorenson, Dan~E. Azagury, and Serena Yeung-Levy.
\newblock Artificial intelligence identifies factors associated with blood loss and surgical experience in cholecystectomy.
\newblock {\em NEJM AI}, 1(2):AIoa2300088, 2024.

\bibitem{knudsen2024clinical}
J~Everett Knudsen, Umar Ghaffar, Runzhuo Ma, and Andrew~J Hung.
\newblock Clinical applications of artificial intelligence in robotic surgery.
\newblock {\em Journal of Robotic Surgery}, 18(1):102, 2024.

\bibitem{lavanchy2023preserving}
Jo{\"e}l~L Lavanchy, Armine Vardazaryan, Pietro Mascagni, Didier Mutter, and Nicolas Padoy.
\newblock Preserving privacy in surgical video analysis using a deep learning classifier to identify out-of-body scenes in endoscopic videos.
\newblock {\em Scientific Reports}, 13(1):9235, 2023.

\bibitem{hegde2024automated}
Shruti~R Hegde, Babak Namazi, Niyenth Iyengar, Sarah Cao, Alexis Desir, Carolina Marques, Heidi Mahnken, Ryan~P Dumas, and Ganesh Sankaranarayanan.
\newblock Automated segmentation of phases, steps, and tasks in laparoscopic cholecystectomy using deep learning.
\newblock {\em Surgical Endoscopy}, 38(1):158--170, 2024.

\bibitem{demir2023deep}
Kubilay~Can Demir, Hannah Schieber, Tobias Weise, Daniel Roth, Matthias May, Andreas Maier, and Seung~Hee Yang.
\newblock Deep learning in surgical workflow analysis: a review of phase and step recognition.
\newblock {\em IEEE Journal of Biomedical and Health Informatics}, 27(11):5405--5417, 2023.

\bibitem{quellec2014real}
Gw{\'e}nol{\'e} Quellec, Mathieu Lamard, B{\'e}atrice Cochener, and Guy Cazuguel.
\newblock Real-time segmentation and recognition of surgical tasks in cataract surgery videos.
\newblock {\em IEEE transactions on medical imaging}, 33(12):2352--2360, 2014.

\bibitem{padoy2008line}
Nicolas Padoy, Tobias Blum, Hubertus Feussner, Marie-Odile Berger, and Nassir Navab.
\newblock On-line recognition of surgical activity for monitoring in the operating room.
\newblock In {\em AAAI}, pages 1718--1724, 2008.

\bibitem{jin2017sv}
Yueming Jin, Qi~Dou, Hao Chen, Lequan Yu, Jing Qin, Chi-Wing Fu, and Pheng-Ann Heng.
\newblock Sv-rcnet: workflow recognition from surgical videos using recurrent convolutional network.
\newblock {\em IEEE transactions on medical imaging}, 37(5):1114--1126, 2017.

\bibitem{valipour2017recurrent}
Sepehr Valipour, Mennatullah Siam, Martin Jagersand, and Nilanjan Ray.
\newblock Recurrent fully convolutional networks for video segmentation.
\newblock In {\em 2017 IEEE Winter Conference on Applications of Computer Vision (WACV)}, pages 29--36. IEEE, 2017.

\bibitem{jalal2021deep}
Nour~Aldeen Jalal, Tamer~Abdulbaki Alshirbaji, Paul~D Docherty, Thomas Neumuth, and Knut Moeller.
\newblock A deep learning framework for recognising surgical phases in laparoscopic videos.
\newblock {\em IFAC-PapersOnLine}, 54(15):334--339, 2021.

\bibitem{he2021db}
Jun-Yan He, Xiao Wu, Zhi-Qi Cheng, Zhaoquan Yuan, and Yu-Gang Jiang.
\newblock Db-lstm: Densely-connected bi-directional lstm for human action recognition.
\newblock {\em Neurocomputing}, 444:319--331, 2021.

\bibitem{sanchez20223dfcnn}
Adrian Sanchez-Caballero, Sergio de~L{\'o}pez-Diz, David Fuentes-Jimenez, Cristina Losada-Guti{\'e}rrez, Marta Marr{\'o}n-Romera, David Casillas-Perez, and Mohammad~Ibrahim Sarker.
\newblock 3dfcnn: Real-time action recognition using 3d deep neural networks with raw depth information.
\newblock {\em Multimedia Tools and Applications}, 81(17):24119--24143, 2022.

\bibitem{hosseini2016alzheimer}
Ehsan Hosseini-Asl, Georgy Gimel'farb, and Ayman El-Baz.
\newblock Alzheimer's disease diagnostics by a deeply supervised adaptable 3d convolutional network.
\newblock {\em arXiv preprint arXiv:1607.00556}, 2016.

\bibitem{huang2018human}
Yi~Huang, Shang-Hong Lai, and Shao-Heng Tai.
\newblock Human action recognition based on temporal pose cnn and multi-dimensional fusion.
\newblock In {\em Proceedings of the European Conference on Computer Vision (ECCV) Workshops}, pages 0--0, 2018.

\bibitem{han2023flatten}
Dongchen Han, Xuran Pan, Yizeng Han, Shiji Song, and Gao Huang.
\newblock Flatten transformer: Vision transformer using focused linear attention.
\newblock In {\em Proceedings of the IEEE/CVF International Conference on Computer Vision}, pages 5961--5971, 2023.

\bibitem{liu2021swin}
Ze~Liu, Yutong Lin, Yue Cao, Han Hu, Yixuan Wei, Zheng Zhang, Stephen Lin, and Baining Guo.
\newblock Swin transformer: Hierarchical vision transformer using shifted windows.
\newblock In {\em Proceedings of the IEEE/CVF international conference on computer vision}, pages 10012--10022, 2021.

\bibitem{arnab2021vivit}
Anurag Arnab, Mostafa Dehghani, Georg Heigold, Chen Sun, Mario Lu{\v{c}}i{\'c}, and Cordelia Schmid.
\newblock Vivit: A video vision transformer.
\newblock In {\em Proceedings of the IEEE/CVF international conference on computer vision}, pages 6836--6846, 2021.

\bibitem{twinanda2016endonet}
Andru~P Twinanda, Sherif Shehata, Didier Mutter, Jacques Marescaux, Michel De~Mathelin, and Nicolas Padoy.
\newblock Endonet: a deep architecture for recognition tasks on laparoscopic videos.
\newblock {\em IEEE transactions on medical imaging}, 36(1):86--97, 2016.

\bibitem{ALHAJJ201924}
Hassan {Al Hajj}, Mathieu Lamard, Pierre-Henri Conze, Soumali Roychowdhury, and Xiaowei~Hu et~al.
\newblock Cataracts: Challenge on automatic tool annotation for cataract surgery.
\newblock {\em Medical Image Analysis}, 52:24--41, 2019.

\bibitem{cadene2016m2cai}
Remi Cadene, Thomas Robert, Nicolas Thome, and Matthieu Cord.
\newblock M2cai workflow challenge: Convolutional neural networks with time smoothing and hidden markov model for video frames classification.
\newblock {\em arXiv preprint arXiv:1610.05541}, 2016.

\bibitem{gao2014jhu}
Y.~Gao, S.~S. Vedula, C.~E. Reiley, N.~Ahmidi, B.~Varadarajan, H.~Liu, L.~Tao, L.~Zappella, B.~Bejar, D.~Yuh, C.~C.~G. Chen, R.~Vidal, S.~Khudanpur, and G.~Hager.
\newblock Jhu-isi gesture and skill assessment working set (jigsaws): A surgical activity dataset for human motion modeling.
\newblock In {\em MICCAI Workshop 2014 M2CAI}, volume~3, 2014.

\bibitem{boreczky1996comparison}
John~S Boreczky and Lawrence~A Rowe.
\newblock Comparison of video shot boundary detection techniques.
\newblock {\em Journal of Electronic Imaging}, 5(2):122--128, 1996.

\bibitem{donahue2015long}
Jeffrey Donahue, Lisa Anne~Hendricks, Sergio Guadarrama, Marcus Rohrbach, Subhashini Venugopalan, Kate Saenko, and Trevor Darrell.
\newblock Long-term recurrent convolutional networks for visual recognition and description.
\newblock In {\em Proceedings of the IEEE conference on computer vision and pattern recognition}, pages 2625--2634, 2015.

\bibitem{savran2023novel}
Rukiye Savran~K{\i}z{\i}ltepe, John~Q Gan, and Juan~Jos{\'e} Escobar.
\newblock A novel keyframe extraction method for video classification using deep neural networks.
\newblock {\em Neural Computing and Applications}, 35(34):24513--24524, 2023.

\bibitem{yang2021fast}
Huimin Yang, Qiuhong Tian, Qiaoli Zhuang, Linye Li, and Qinglong Liang.
\newblock Fast and robust key frame extraction method for gesture video based on high-level feature representation.
\newblock {\em Signal, Image and Video Processing}, 15:617--626, 2021.

\bibitem{zhong2020key}
Qi~Zhong, Yuan Zhang, Jinguo Zhang, Kaixuan Shi, Yang Yu, and Chang Liu.
\newblock Key frame extraction algorithm of motion video based on priori.
\newblock {\em IEEE Access}, 8:174424--174436, 2020.

\bibitem{cucchiara2003detecting}
Rita Cucchiara, Costantino Grana, Massimo Piccardi, and Andrea Prati.
\newblock Detecting moving objects, ghosts, and shadows in video streams.
\newblock {\em IEEE transactions on pattern analysis and machine intelligence}, 25(10):1337--1342, 2003.

\bibitem{redmon2016you}
Joseph Redmon, Santosh Divvala, Ross Girshick, and Ali Farhadi.
\newblock You only look once: Unified, real-time object detection.
\newblock In {\em Proceedings of the IEEE conference on computer vision and pattern recognition}, pages 779--788, 2016.

\bibitem{girshick2014rich}
Ross Girshick, Jeff Donahue, Trevor Darrell, and Jitendra Malik.
\newblock Rich feature hierarchies for accurate object detection and semantic segmentation.
\newblock In {\em Proceedings of the IEEE conference on computer vision and pattern recognition}, pages 580--587, 2014.

\bibitem{lin2014microsoft}
Tsung-Yi Lin, Michael Maire, Serge Belongie, James Hays, Pietro Perona, Deva Ramanan, Piotr Doll{\'a}r, and C~Lawrence Zitnick.
\newblock Microsoft coco: Common objects in context.
\newblock In {\em Computer Vision--ECCV 2014: 13th European Conference, Zurich, Switzerland, September 6-12, 2014, Proceedings, Part V 13}, pages 740--755. Springer, 2014.

\bibitem{deng2009imagenet}
Jia Deng, Wei Dong, Richard Socher, Li-Jia Li, Kai Li, and Li~Fei-Fei.
\newblock Imagenet: A large-scale hierarchical image database.
\newblock In {\em 2009 IEEE conference on computer vision and pattern recognition}, pages 248--255. Ieee, 2009.

\bibitem{russell2008labelme}
Bryan~C Russell, Antonio Torralba, Kevin~P Murphy, and William~T Freeman.
\newblock Labelme: a database and web-based tool for image annotation.
\newblock {\em International journal of computer vision}, 77:157--173, 2008.

\bibitem{shekhar2024machine}
Madhav~Khairnar Shekhar, Nguyen Huu~Phong, Desir Alexis, Holcomb Carla, J.~Scott Daniel, and Sankaranarayanan Ganesh.
\newblock Machine learning-based automated assessment of intracorporeal suturing in laparoscopic fundoplication.
\newblock {\em arXiv preprint arXiv:2412.16195}, 2024.

\bibitem{feichtenhofer2020x3d}
Christoph Feichtenhofer.
\newblock X3d: Expanding architectures for efficient video recognition.
\newblock In {\em Proceedings of the IEEE/CVF conference on computer vision and pattern recognition}, pages 203--213, 2020.

\bibitem{he2016deep}
Kaiming He, Xiangyu Zhang, Shaoqing Ren, and Jian Sun.
\newblock Deep residual learning for image recognition.
\newblock In {\em Proceedings of the IEEE conference on computer vision and pattern recognition}, pages 770--778, 2016.

\bibitem{funke2023metrics}
Isabel Funke, Dominik Rivoir, and Stefanie Speidel.
\newblock Metrics matter in surgical phase recognition.
\newblock {\em arXiv preprint arXiv:2305.13961}, 2023.

\bibitem{guedon2021deep}
Annetje~CP Gu{\'e}don, Senna~EP Meij, Karim~NMMH Osman, Helena~A Kloosterman, Karlijn~J van Stralen, Matthijs~CM Grimbergen, Quirijn~AJ Eijsbouts, John~J van~den Dobbelsteen, and Andru~P Twinanda.
\newblock Deep learning for surgical phase recognition using endoscopic videos.
\newblock {\em Surgical endoscopy}, 35:6150--6157, 2021.

\bibitem{touma2022development}
Samir Touma, Fares Antaki, and Renaud Duval.
\newblock Development of a code-free machine learning model for the classification of cataract surgery phases.
\newblock {\em Scientific Reports}, 12(1):2398, 2022.

\bibitem{meireles2021sages}
Ozanan~R Meireles, Guy Rosman, Maria~S Altieri, Lawrence Carin, Gregory Hager, Amin Madani, Nicolas Padoy, Carla~M Pugh, Patricia Sylla, Thomas~M Ward, et~al.
\newblock Sages consensus recommendations on an annotation framework for surgical video.
\newblock {\em Surgical endoscopy}, 35(9):4918--4929, 2021.

\bibitem{golany2022artificial}
Tomer Golany, Amit Aides, Daniel Freedman, Nadav Rabani, Yun Liu, Ehud Rivlin, Greg~S Corrado, Yossi Matias, Wisam Khoury, Hanoch Kashtan, et~al.
\newblock Artificial intelligence for phase recognition in complex laparoscopic cholecystectomy.
\newblock {\em Surgical Endoscopy}, 36(12):9215--9223, 2022.

\bibitem{garrow2021machine}
Carly~R Garrow, Karl-Friedrich Kowalewski, Linhong Li, Martin Wagner, Mona~W Schmidt, Sandy Engelhardt, Daniel~A Hashimoto, Hannes~G Kenngott, Sebastian Bodenstedt, Stefanie Speidel, et~al.
\newblock Machine learning for surgical phase recognition: a systematic review.
\newblock {\em Annals of surgery}, 273(4):684--693, 2021.

\bibitem{tran2017phase}
Dinh~Tuan Tran, Ryuhei Sakurai, Hirotake Yamazoe, Joo-Ho Lee, et~al.
\newblock Phase segmentation methods for an automatic surgical workflow analysis.
\newblock {\em International journal of biomedical imaging}, 2017, 2017.

\bibitem{pydimarry2024evaluating}
Sai~Abhinav Pydimarry, Shekhar~Madhav Khairnar, Sofia~Garces Palacios, Ganesh Sankaranarayanan, Darian Hoagland, Dmitry Nepomnayshy, and Huu~Phong Nguyen.
\newblock Evaluating model performance with hard-swish activation function adjustments.
\newblock {\em RECPAD}, 2024.

\bibitem{farneback2003two}
Gunnar Farneb{\"a}ck.
\newblock Two-frame motion estimation based on polynomial expansion.
\newblock In {\em Image Analysis: 13th Scandinavian Conference, SCIA 2003 Halmstad, Sweden, June 29--July 2, 2003 Proceedings 13}, pages 363--370. Springer, 2003.

\bibitem{vining2021robotic}
Charles~C Vining, Kristine Kuchta, Yaniv Berger, Pierce Paterakos, Darryl Schuitevoerder, Kevin~K Roggin, Mark~S Talamonti, and Melissa~E Hogg.
\newblock Robotic pancreaticoduodenectomy decreases the risk of clinically relevant post-operative pancreatic fistula: a propensity score matched nsqip analysis.
\newblock {\em HPB}, 23(3):367--378, 2021.

\bibitem{finks2011trends}
John~F. Finks, Nicholas~H. Osborne, and John~D. Birkmeyer.
\newblock Trends in hospital volume and operative mortality for high-risk surgery.
\newblock {\em New England Journal of Medicine}, 364(22):2128--2137, 2011.

\bibitem{crist1987improved}
DAVID~W Crist, JAMES~V Sitzmann, and JOHN~L Cameron.
\newblock Improved hospital morbidity, mortality, and survival after the whipple procedure.
\newblock {\em Annals of surgery}, 206(3):358, 1987.

\bibitem{olsen2024surgical}
Rikke~Groth Olsen, Morten Bo~S{\o}ndergaard Svendsen, Martin~G Tolsgaard, Lars Konge, Andreas R{\o}der, and Flemming Bjerrum.
\newblock Surgical gestures can be used to assess surgical competence in robot-assisted surgery: A validity investigating study of simulated rarp.
\newblock {\em Journal of Robotic Surgery}, 18(1):47, 2024.

\bibitem{czempiel2020tecno}
Tobias Czempiel, Magdalini Paschali, and Matthias et~al. Keicher.
\newblock Tecno: Surgical phase recognition with multi-stage temporal convolutional networks.
\newblock In {\em Medical Image Computing and Computer Assisted Intervention -- MICCAI 2020}, pages 343--352, Cham, 2020. Springer International Publishing.

\bibitem{liu2025lovit}
Yang Liu, Maxence Boels, Luis~C Garcia-Peraza-Herrera, Tom Vercauteren, Prokar Dasgupta, Alejandro Granados, and Sebastien Ourselin.
\newblock Lovit: Long video transformer for surgical phase recognition.
\newblock {\em Medical Image Analysis}, 99:103366, 2025.

\bibitem{jin2020multi}
Yueming Jin, Huaxia Li, Qi~Dou, Hao Chen, Jing Qin, Chi-Wing Fu, and Pheng-Ann Heng.
\newblock Multi-task recurrent convolutional network with correlation loss for surgical video analysis.
\newblock {\em Medical image analysis}, 59:101572, 2020.

\bibitem{luo2024surgplan}
Xingjian Luo, You Pang, Zhen Chen, Jinlin Wu, Zongmin Zhang, Zhen Lei, and Hongbin Liu.
\newblock Surgplan: Surgical phase localization network for phase recognition.
\newblock In {\em 2024 IEEE International Symposium on Biomedical Imaging (ISBI)}, pages 1--5. IEEE, 2024.

\bibitem{holm2023dynamic}
Felix Holm, Ghazal Ghazaei, Tobias Czempiel, Ege {\"O}zsoy, Stefan Saur, and Nassir Navab.
\newblock Dynamic scene graph representation for surgical video.
\newblock In {\em Proceedings of the IEEE/CVF international conference on computer vision}, pages 81--87, 2023.

\bibitem{tao2023last}
Rong Tao, Xiaoyang Zou, and Guoyan Zheng.
\newblock Last: Latent space-constrained transformers for automatic surgical phase recognition and tool presence detection.
\newblock {\em IEEE Transactions on Medical Imaging}, 42(11):3256--3268, 2023.

\bibitem{loukas2018keyframe}
Constantinos Loukas, Christos Varytimidis, Konstantinos Rapantzikos, and Meletios~A Kanakis.
\newblock Keyframe extraction from laparoscopic videos based on visual saliency detection.
\newblock {\em Computer methods and programs in biomedicine}, 165:13--23, 2018.

\bibitem{le1991mpeg}
Didier Le~Gall.
\newblock Mpeg: A video compression standard for multimedia applications.
\newblock {\em Communications of the ACM}, 34(4):46--58, 1991.

\bibitem{nguyen2019advanced}
Huu~Phong Nguyen and Bernardete Ribeiro.
\newblock Advanced capsule networks via context awareness.
\newblock In {\em Artificial Neural Networks and Machine Learning--ICANN 2019: Theoretical Neural Computation: 28th International Conference on Artificial Neural Networks, Munich, Germany, September 17--19, 2019, Proceedings, Part I 28}, pages 166--177. Springer, 2019.

\bibitem{phong2019improvement}
Huu~Phong Nguyen and Bernardete Ribeiro.
\newblock An improvement for capsule networks using depthwise separable convolution.
\newblock In {\em Iberian conference on pattern recognition and image analysis}, pages 521--530. Springer, 2019.

\bibitem{phong2018action}
Huu~Phong Nguyen and Bernardete Ribeiro.
\newblock Action recognition for american sign language.
\newblock {\em RECPAD}, 2018.

\end{thebibliography}

% \newpage
\clearpage
\begin{IEEEbiography}
[{\includegraphics[width=1in,height=1.2in,clip,keepaspectratio]{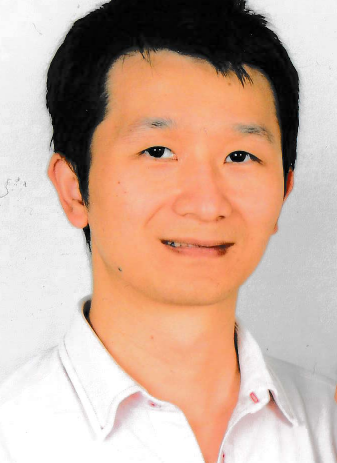}}]%
{Huu Phong Nguyen}
received a BSc in Physics from Vietnam National University, Hanoi, an MSc in Information Technology from Shinawatra University, Thailand, and a PhD degree from the University of Coimbra, Portugal. He was a member of the Adaptive Computation group at the Centre of Informatics and Systems at the University of Coimbra (CISUC). Currently, he is pursuing a PostDoc at the University of Texas Medical Center, USA. His research interests include medical imaging, image classification, action recognition, pattern recognition, and deep learning.
\end{IEEEbiography}

\begin{IEEEbiography}
[{\includegraphics[width=1in,height=1.2in,clip,keepaspectratio]{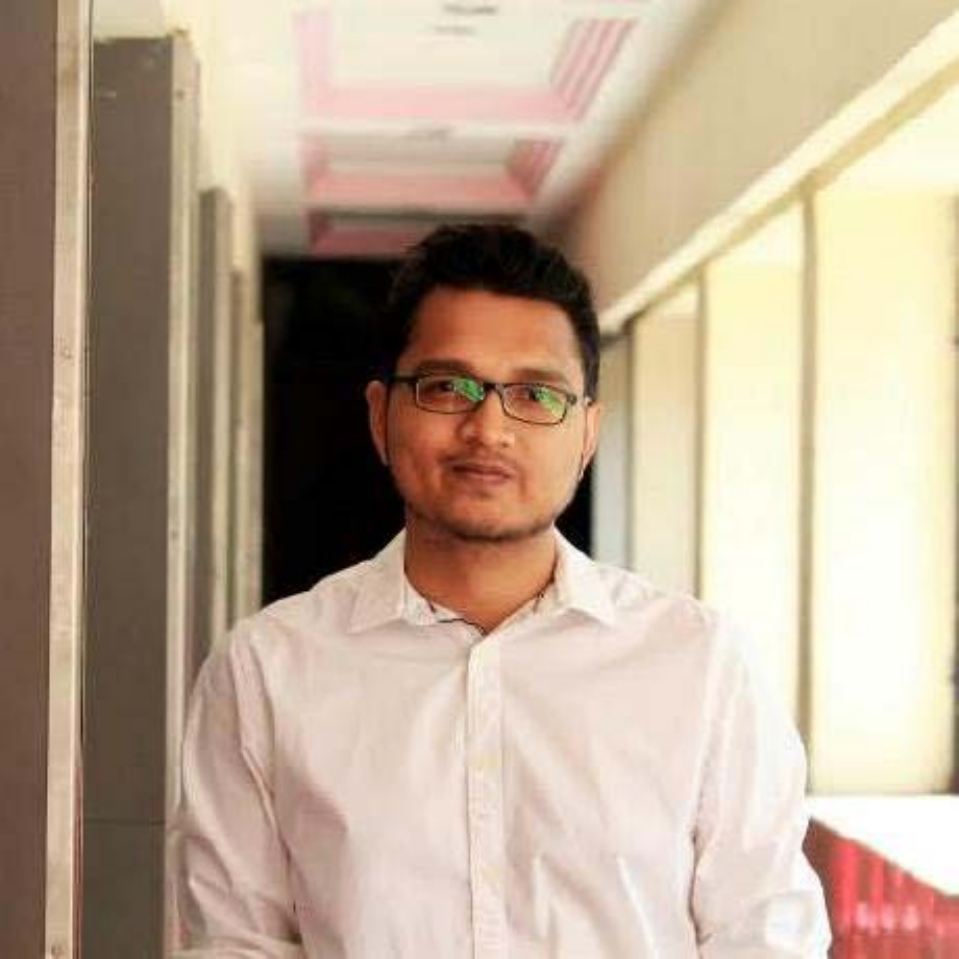}}]%
{Shekhar Madhav Khairnar} 
is currently working as a Data Scientist at UT Southwestern Medical Center, focusing on developing AI models, particularly in computer vision domain. He earned a Bachelor of Engineering (BE) in Electrical from the University of Pune, India, and a Master of Science (MS) in Industrial Engineering from the University of Texas at Arlington, where he received awards including the GTA/GRA and G.T. Stevens Endowed Scholarship. His research areas include computer vision, covering image classification, segmentation, object detection, and tracking, and broader fields such as machine learning, natural language processing (NLP), large language models (LLMs), and deep learning.
\end{IEEEbiography}

\begin{IEEEbiography}
[{\includegraphics[width=1in,height=1.2in,clip,keepaspectratio]{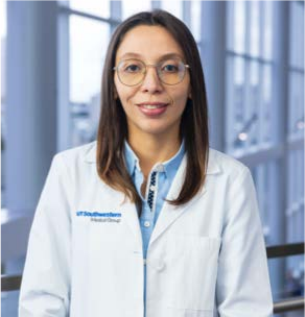}}]%
{Sofia Garces Palacios}
is a Research Associate at the Artificial Intelligent and Medical Simulation (AIMS) Laboratory within the Department of Surgery at UT Southwestern Medical Center. Dr. Sofia holds a Doctor of Medicine degree from the University of Manizales in Colombia and is pursuing a career in General Surgery and minimal invasive fellowship. Her research interests are centered on surgical education, minimally invasive surgical techniques, and global surgery. 
\end{IEEEbiography}

\begin{IEEEbiography}
[{\includegraphics[width=1in,height=1.2in,clip,keepaspectratio]{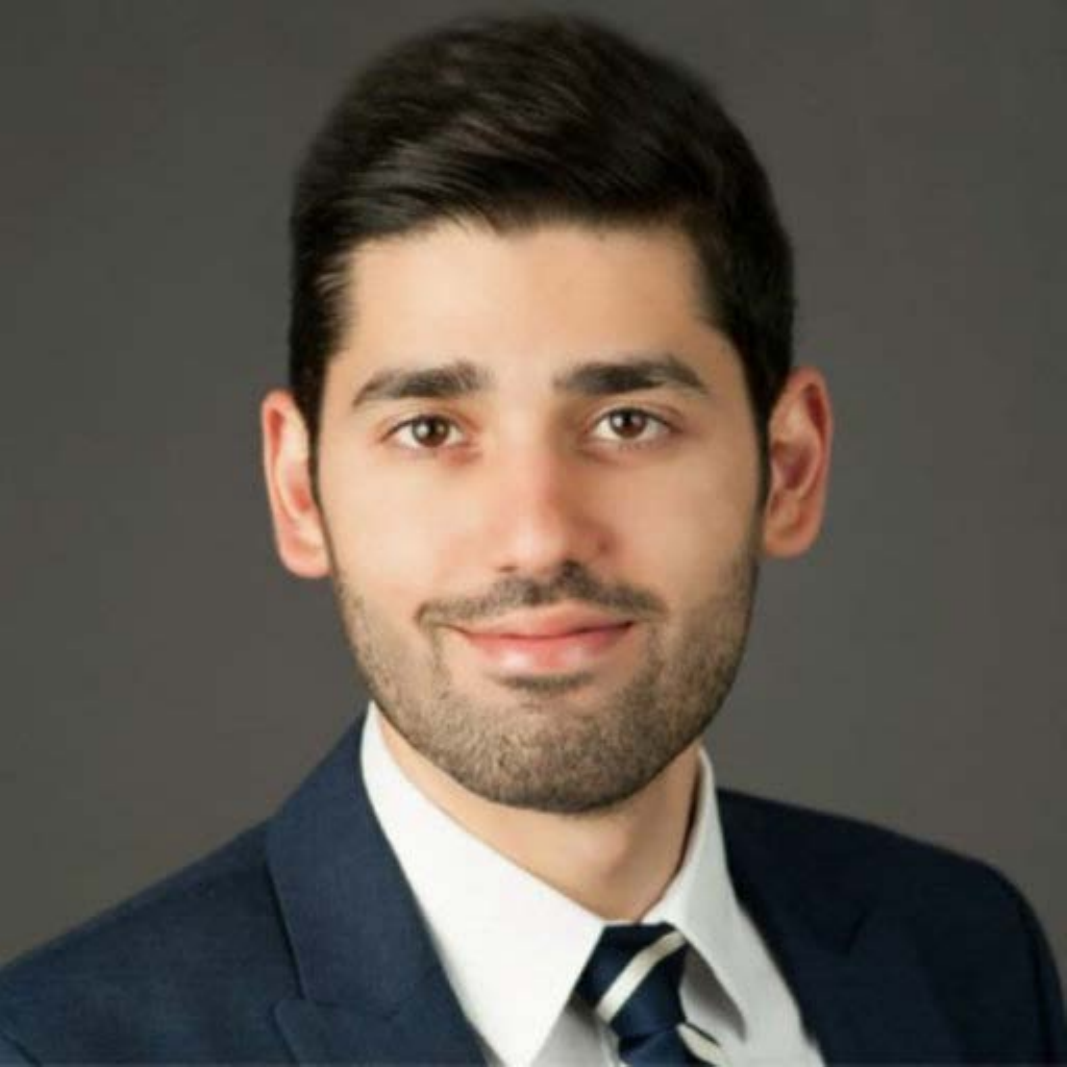}}]%
{Amr Al-Abbas} is a General Surgery Resident at UT Southwestern Medical Center and Parkland Hospital. Dr Al Abbas will start his fellowship in Surgical Oncology at UPMC in 2025. He also served as a Postdoctoral Research Fellow at UT Southwestern, focusing on artificial intelligence for video-based assessment in robotic pancreatic surgery. Dr. Al Abbas also completed postdoctoral research in surgical oncology at UPMC and the University of Pittsburgh, where he conducted clinical outcomes research. Dr. Al Abbas holds a Doctor of Medicine degree from the American University of Beirut.  Dr. Amr Al Abbas is interested in surgical oncology robotic, surgery surgical, education video-based assessment, machine learning.
\end{IEEEbiography}

\begin{IEEEbiography}
[{\includegraphics[width=1in,height=1.2in,clip,keepaspectratio]{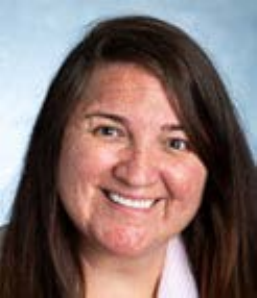}}]%
{Melissa E. Hogg}
is the Division Chief of GI and General Surgery at Northshore University Hospital and Director of Advanced Robotic Training for the Grainger Center for Simulation and Innovation. She did her medical school and residency training at Northwestern University in Chicago, IL, and her fellowship and masters training in Medical Education and Clinical Research at the University of Pittsburgh, PA. She is faculty of the University of Chicago Pritzker School of Medicine and serves as Associate Program Director of the General Surgery Residency. Dr. Hogg’s clinical practice specializes in abdominal cancers and has a focus in minimally invasive gastrointestinal surgery and performs robotic hepatobiliary surgery.
\end{IEEEbiography}

\begin{IEEEbiography}
[{\includegraphics[width=1in,height=1.2in,clip,keepaspectratio]{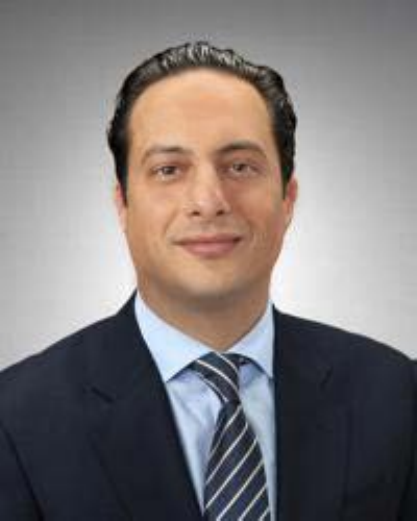}}]%
{Amer H. Zureikat}
is the Chief of the Division of Gastrointestinal Surgical Oncology and Associate Professor of Surgery at the University of Pittsburgh School of Medicine. He also is the Vice Chair of Surgery, Surgical Oncology at the UPMC Department of Surgery and Chief of Surgical Oncology at UPMC Hillman Cancer Center. Dr. Zureikat is board-certified in general surgery. He received his medical degree from the Royal College of Surgeons in Dublin, Ireland, and completed his residency in general surgery at the University of Chicago Medical Center and his fellowship in surgical oncology at UPMC. His clinical and research interests focus on comparative effectiveness studies that evaluate outcomes of minimally invasive and robotic approaches to pancreatic and gastrointestinal cancers.
\end{IEEEbiography}

\begin{IEEEbiography}
[{\includegraphics[width=1in,height=1.2in,clip,keepaspectratio]{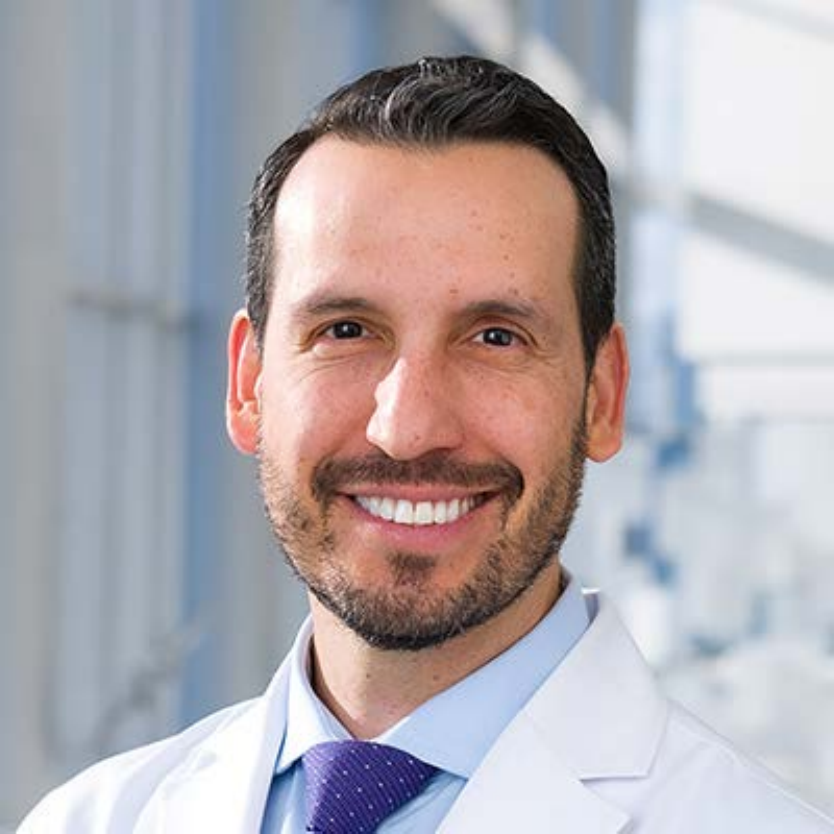}}]%
{Patricio Polanco}
is an Associate Professor in the Division of Surgical Oncology in the Department of Surgery at UT Southwestern Medical Center. He serves as the Director of the Robotic Surgery Training Program at UT Southwestern as well as Co-director for the Pancreatic Cancer Program and the Pancreatic Cancer Prevention Clinic. He also leads the Peritoneal Surface Malignancies and HIPEC (Hyperthermic Intraperitoneal Chemotherapy) Program at UT Southwestern's Harold C. Simmons Comprehensive Cancer Center. A double board-certified surgeon by the American Board of Surgery (general surgery and complex general surgical oncology), Dr. Polanco joined the UT Southwestern faculty in 2014. Dr. Polanco graduated as valedictorian of Universidad de San Martin de Porres Medical School. After completing his surgical training in Lima, Peru, he moved to the United States, where he completed a postdoctoral research fellowship and a general surgery residency at the University of Pittsburgh. His clinical interests and expertise include pancreatic cancer, liver cancer, colorectal cancer, neuroendocrine tumors, minimally invasive techniques, robotic surgery, HIPEC surgery, and hepatic artery infusion pump surgery, among other complex gastrointestinal diseases and procedures. Dr. Polanco's research interests include health services research, disparities, and surgical outcomes in hepato-pancreato-biliary (HPB) malignancies and pancreatic cancer. 
\end{IEEEbiography}

\begin{IEEEbiography}
[{\includegraphics[width=1in,height=1.2in,clip,keepaspectratio]{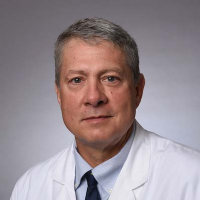}}]%
{Herbert J. Zeh III} 
is Professor and Chair of the Department of Surgery at UT Southwestern Medical Center, where he holds the Hall and Mary Lucile Shannon Distinguished Chair in Surgery. A leader in pancreatic diseases and cancer, Dr. Zeh has over 20 years of surgical experience. He earned his medical degree from the University of Pittsburgh and completed his residency and fellowship at The Johns Hopkins Hospital. Before joining UT Southwestern in 2018, he served as the Watson Family Professor of Surgery at the University of Pittsburgh and co-directed the UPMC Pancreatic Cancer Center. A prolific author with over 200 peer-reviewed publications, Dr. Zeh’s research focuses on novel treatments for pancreatic cancer and robotic surgery innovations. He is a member of numerous prestigious surgical organizations. 
\end{IEEEbiography}

\begin{IEEEbiography}
[{\includegraphics[width=1in,height=1.2in,clip,keepaspectratio]{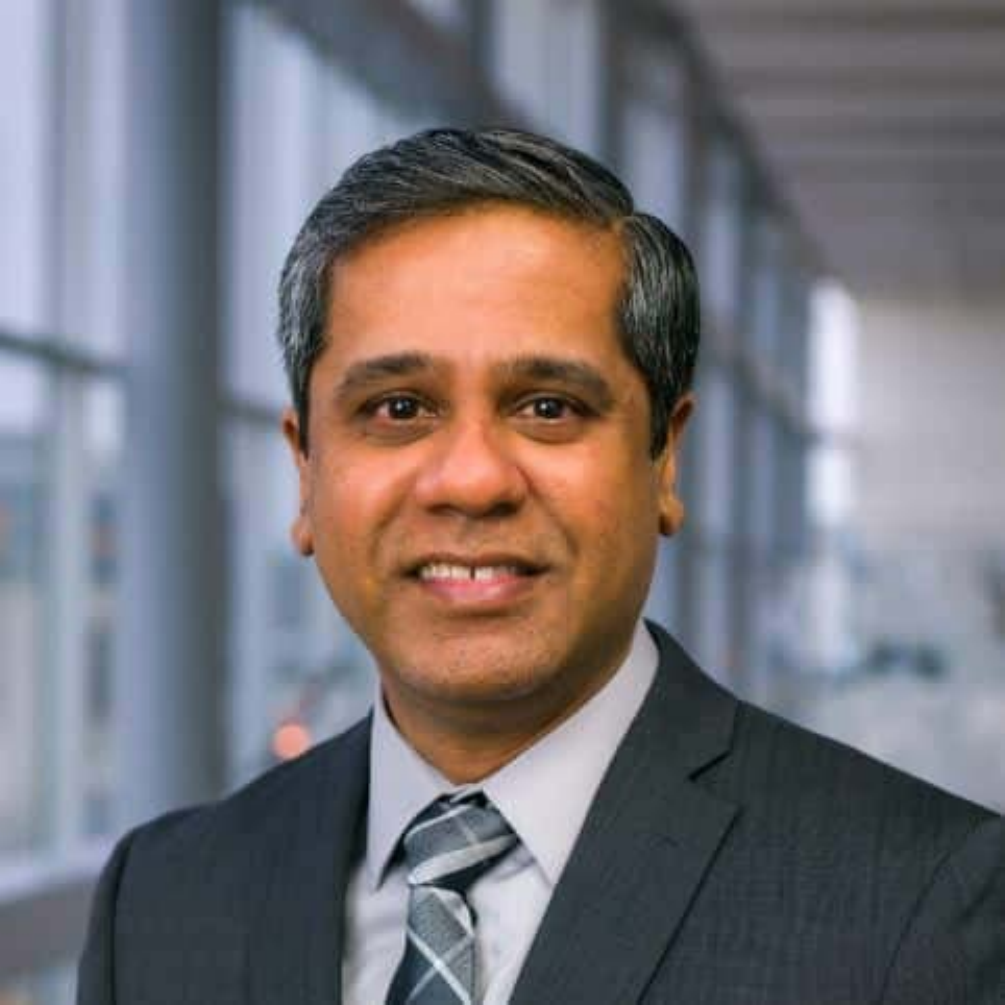}}]%
{Ganesh Sankaranarayanan}
is an Associate Professor in the Departments of Surgery and Biomedical Engineering at UT Southwestern Medical Center. He co-directs the Center for Assessment of Surgical Proficiency, where his work focuses on surgical simulation, virtual reality, and artificial intelligence applications in surgical training and assessment. He earned his Ph.D. in Electrical Engineering from the University of Washington and completed postdoctoral research at Rensselaer Polytechnic Institute, concentrating on virtual surgery simulation and medical robotics. His career includes roles as Assistant Director at Baylor University Medical Center’s Center for Evidence Based Simulation and faculty positions at Texas A\&M College of Medicine and the University of Texas at Arlington. His research has been supported by the NIH and industry partners, leading to innovations such as AI-driven surgical video analysis, haptic-enabled simulators, and VR-based training tools. He is a member of surgical societies that includes the ACS, SAGES, ASE and technical societies that includes the IEEE, IEEE Robotics and Automation Society and IEEE EMBS.\end{IEEEbiography}
% \vfill
%
\EOD

\end{document}